\documentclass{article}

    \usepackage[preprint, nonatbib]{neurips_2020}

\usepackage[utf8]{inputenc} 
\usepackage[T1]{fontenc}    
\usepackage{hyperref}       
\usepackage{url}            
\usepackage{booktabs}       
\usepackage{amsfonts}       
\usepackage{nicefrac}       
\usepackage{microtype}      

\usepackage{microtype}
\usepackage{graphicx}
\usepackage{subfigure}
\usepackage{booktabs} 
\usepackage{amssymb,amsmath, amsthm}

\usepackage{algorithm, algorithmic}

\newcommand{\bE}{{\mathbb{E}}}

\newcommand{\bN}{{\mathbb{N}}}
\newcommand{\bR}{{\mathbb{R}}}

\newcommand{\cA}{\mathcal{A}}
\newcommand{\cC}{\mathcal{C}}

\newcommand{\cE}{\mathcal{E}}

\newcommand{\cH}{\mathcal{H}}

\newcommand{\cN}{\mathcal{N}}
\newcommand{\cQ}{\mathcal{Q}}

\newcommand{\cT}{\mathcal{T}}

\newcommand{\beq}{\begin{equation}}
\newcommand{\eeq}{\end{equation}}

\newtheorem{theorem}{Theorem}

\DeclareMathOperator*{\argmin}{arg\,min}

\title{Projected Stein Variational Gradient Descent}

\author{%
  Peng Chen $\quad \quad \quad $Omar Ghattas \\
	Oden Institute for Computational Engineering and Sciences \\
	The University of Texas at Austin \\
	Austin, TX 78712.\\
  \texttt{\{peng, omar\}@oden.utexas.edu} 
}

\begin{document}

\maketitle

\begin{abstract}
The curse of dimensionality is a longstanding challenge in Bayesian
inference in high dimensions. In this work, we propose a {projected
  Stein variational gradient descent} (pSVGD) method to overcome this
challenge by exploiting the fundamental property of intrinsic low
dimensionality of the data informed subspace stemming from
ill-posedness of such problems. We adaptively construct the subspace
using a gradient information matrix of the log-likelihood, and apply
pSVGD to the much lower-dimensional coefficients of the parameter
projection. The method is demonstrated to be more accurate and
efficient than SVGD. It is also shown to be more scalable with respect
to the number of parameters, samples, data points, and processor cores
via experiments with parameters dimensions ranging from the
hundreds to the tens of thousands. 
\end{abstract}

\section{Introduction}
\label{sec:introduction}
Given observation data for a system with unknown parameters, Bayesian inference provides an optimal probability framework for learning the parameters by updating their prior distribution to a posterior distribution. However, many conventional methods for solving high-dimensional Bayesian inference problems face the curse of dimensionality, i.e., the computational complexity grows rapidly, often exponentially, with respect to (w.r.t.) the number of parameters. 
To address the curse of dimensionality, the intrinsic properties of the posterior distribution, such as its smoothness, sparsity, and intrinsic low-dimensionality, have been exploited to reduce the parameter correlation and develop efficient methods whose complexity grows slowly or remains the same with increasing dimension. By exploiting the geometry of the log-likelihood function, accelerated Markov chain Monte Carlo (MCMC) methods have been developed to reduce the sample correlation or increase effective sample size independent of the dimension \cite{GirolamiCalderhead11, MartinWilcoxBursteddeEtAl12, PetraMartinStadlerEtAl14, ConstantineKentBui-Thanh16, CuiLawMarzouk16, BESKOS2017327}.
Nevertheless, these random and essentially serial sampling methods remain prohibitive for large-scale inference problems with expensive likelihoods. Deterministic methods using sparse quadratures \cite{SchwabStuart12, SchillingsSchwab2013, ChenSchwab2015, ChenSchwab2016} were shown to converge rapidly with dimension-independent rates for problems with smooth and sparse posteriors. However, for posteriors lacking smoothness or sparsity, the convergence deteriorates significantly, despite incorporation of Hessian-based transformations \cite{SchillingsSchwab16, ChenVillaGhattas2017}. 

Transport-based variational inference is another type of deterministic method that seeks a transport map in a function space (represented by, e.g., polynomials, kernels, or neural networks) that pushes the prior to the posterior by minimizing the difference between the transported prior and the posterior, measured in, e.g., Kullback--Leibler divergence \cite{MarzoukMoselhyParnoEtAl16, LiuWang16, BleiKucukelbirMcAuliffe17, BigoniZahmSpantiniEtAl19, DetommasoKruseArdizzoneEtAl19}. In particular, kernel-based Stein variational methods, using gradient-based (SVGD) \cite{LiuWang16, ChenMackeyGorhamEtAl18, LiuZhu18} and Hessian-based (SVN) \cite{DetommasoCuiMarzoukEtAl18, WangLi20} optimization methods, are shown to achieve fast convergence in relatively low dimensions.
Nonetheless, the convergence and accuracy of these methods deteriorates in high dimensions due to the curse of dimensionality in kernel representation. This can be partially addressed by a localized SVGD on Markov blankets, which relies on conditional independence of the target distribution \cite{ZhuoLiuShiEtAl18, WangZengLiu18}, or by parameter projection for dimension reduction in pSVN \cite{ChenWuChenEtAl19} and lazy maps \cite{BigoniZahmSpantiniEtAl19}.

\textbf{Contributions}: 
Here, we propose, analyze, and apply a projected SVGD method to tackle the curse of dimensionality for high-dimensional nonlinear Bayesian inference problems, which relies on the fundamental property that the posterior effectively differs from the prior only in a low-dimensional subspace of high-dimensional parameters, see \cite{Bui-ThanhGhattasMartinEtAl2013, SpantiniSolonenCuiEtAl15, IsaacPetraStadlerEtAl15, CuiLawMarzouk16, ChenVillaGhattas2017, ChenVillaGhattas18, ChenGhattas19, BigoniZahmSpantiniEtAl19, ChenWuChenEtAl19} and references therein. Specifically, our contributions are: (1) we propose dimension reduction by projecting the parameters to a low-dimensional subspace constructed using the gradient of the log-likelihood, and push the prior samples of the projection coefficients to their posterior by pSVGD; (2) we prove the equivalence of the projected transport in the coefficient space and the transport in the projected parameter space; 
(3) we propose adaptive and parallel algorithms to efficiently approximate the optimal profile function and the gradient information matrix for the construction of the subspace; and (4) we demonstrate the accuracy (compared to SVGD) and scalability of pSVGD w.r.t.\ the number of parameters, samples, data points, and processor cores by classical high-dimensional Bayesian inference problems.

The major differences of this work compared to pSVN \cite{ChenWuChenEtAl19} are:
(1) pSVGD uses only gradient information of the log-likelihood, 
which is available for many models, 
while pSVN requires Hessian information, which is challenging for complex models and codes in practical applications; (2) the upper bound for the projection error w.r.t.\ the posterior is sharper than that for pSVN; (3) we prove the equivalence of the projected transport for the coefficient and the transport for the projected parameters; (4) we also test new benchmark examples and investigate the convergence of pSVGD w.r.t.\ the number of parameters and the scalability of pSVGD w.r.t.\ the number of data points.


\section{Preliminaries}
\label{sec:SVGD}
Let $x \in \bR^d$ denote a random parameter of dimension $d \in \bN$, which has a continuous prior density $p_0: \bR^d \to \bR$. Let $y = \{y_i\}_{i=1}^s$ denote a set of i.i.d.\ observation data. Let $f(x) := \prod_{i=1}^s p(y_i|x)$ denote, up to a multiplicative constant, a continuous likelihood of $y$ at given $x$. Then the posterior density of parameter $x$ conditioned on data $y$, denoted as $p(\cdot): \bR^d \to \bR$, is given by Bayes' rule as 
\beq\label{eq:Bayes}
p(x) = \frac{1}{Z} f(x) p_0(x), 
\eeq
where $Z$ is the normalization constant defined as
\beq
Z = \int_{\bR^d} f(x) p_0(x) dx,
\eeq
whose computation is typically intractable, especially for a large $d$. The central task of Bayesian inference is to draw samples of parameter $x$ from its posterior with density $p$, and compute some statistical quantity of interest, e.g., the mean and variance of the parameter $x$ or some function of $x$. 

SVGD is one type of variational inference method that seeks an approximation of the posterior density $p$ by a function $q^*$ in a predefined function set $\cQ$, which is realized by minimizing the Kullback--Leibler (KL) divergence that measures the difference between two densities, i.e., 
\beq
q^* = \argmin_{q \in \cQ} D_{\text{KL}} (q | p),
\eeq
where $D_{\text{KL}}(q|p) = \bE_{x \sim q}[\log(q/p)]$, i.e., the average of $\log(q/p)$ with respect to the density $q$, which vanishes when $q = p$. In particular, a transport based function set is considered as $\cQ = \{T_\sharp p_0: T \in \cT\}$, where $T_\sharp$ is a pushforward map that pushes the prior density to a new density $q := T_\sharp p_0$ through an invertible transport map $T(\cdot):\bR^d \to \bR^d$ in a space $\cT$.
Let $T$ be given by
\beq\label{eq:transport}
T(x) = x + \epsilon \phi(x),
\eeq
where $\phi : \bR^d \to \bR^d$ is a differentiable perturbation map w.r.t.\ $x$, and $\epsilon > 0$ is small enough so that $T$ is invertible. It is shown in \cite{LiuWang16} that 
\beq\label{eq:directional-gradient}
\nabla_\epsilon D_{\text{KL}}(T_\sharp p_0 | p) \big|_{\epsilon = 0} = - \bE_{x\sim p_0} [\text{trace}(\cA_p \phi(x))],
\eeq
where $\cA_p$ is the Stein operator given by 
\beq
\cA_p \phi(x) = \nabla_x \log p(x) \phi(x)^T + \nabla_x \phi(x). 
\eeq
Based on this result, a practical SVGD algorithm was developed in \cite{LiuWang16} by choosing the space $\cT = (\cH_d)^d = \cH_d \times \cdots \times \cH_d$, a tensor product of a reproducing kernel Hilbert space (RKHS) $\cH_d$ with kernel $k(\cdot, \cdot): \bR^d \times \bR^d \to \bR$. SVGD updates samples $x^0_1, \dots, x^0_N$ drawn from the prior $p_0$ as
\beq
x_m^{\ell+1} = x_m^\ell + \epsilon_l \hat{\phi}_{\ell}^*(x_m^\ell), \quad m = 1, \dots, N, \ell = 0, 1, \dots  
\eeq
where $\epsilon_l$ is a step size or learning rate, and $\hat{\phi}_{\ell}^*(x_m^\ell) $ is an approximate steepest direction given by
\beq\label{eq:gradient-saa}
\hat{\phi}_{\ell}^*(x_m^\ell) = \frac{1}{N} \sum_{n=1}^N \nabla_{x_n^\ell} \log p(x_n^\ell) k(x_n^\ell, x_m^\ell) + \nabla_{x_n^\ell} k(x_n^\ell, x_m^\ell).
\eeq
The kernel $k$ plays a critical role in pushing the samples to the posterior. One choice is Gaussian
\beq\label{eq:x-kernel}
k(x,x') = \exp\left(- \frac{||x-x'||^2_2}{h}\right),
\eeq
where $h$ is the bandwidth, e.g., $h = \text{med}^2/\log(N)$ with $\text{med}$ representing the median of sample distances \cite{LiuWang16}. 
However, it is known that the kernel suffers from the \emph{curse of dimensionality} for large $d$ \cite{RamdasReddiPoczosEtAl15, ZhuoLiuShiEtAl18, WangZengLiu18}, which leads to samples not representative of the posterior, as observed in \cite{ZhuoLiuShiEtAl18, WangZengLiu18}.

\section{Projected Stein Variational Gradient Descent}
\label{sec:pSVGD}

To tackle the curse of dimensionality of SVGD, we exploit one fundamental property of many high-dimensional Bayesian inference problems --- the posterior only effectively differs from the prior in a relatively low-dimensional subspace due to the ill-posedness or over-parametrization of the inference problems, see many examples and some proofs in, e.g., \cite{BashirWillcoxGhattasEtAl08, Bui-ThanhGhattas2012, Bui-ThanhGhattasMartinEtAl2013, SpantiniSolonenCuiEtAl15, IsaacPetraStadlerEtAl15, CuiLawMarzouk16, ChenVillaGhattas2017, ChenVillaGhattas18, ChenGhattas19, BigoniZahmSpantiniEtAl19, ChenWuChenEtAl19}. 

\subsection{Dimension reduction by projection}
\label{sec:dimension-reduction}

By $H \in \bR^{d\times d}$ we denote a gradient information matrix, which is defined as the average of the outer product of the gradient of the log-likelihood w.r.t.\ the posterior, i.e.,  
\beq\label{eq:grad-grad-T}
H = \int_{\bR^d} (\nabla_x \log f(x)) (\nabla_x \log f(x))^T p(x) dx. 
\eeq
By $(\lambda_i, \psi_i)_{i=1}^r$ we denote the dominant eigenpairs of $(H, \Gamma)$, with $\Gamma$ representing the covariance of the parameter $x$ w.r.t.\ its prior, i.e., $(\lambda_i, \psi_i)_{i=1}^r$ correspond to the $r$ largest eigenvalues $\lambda_1 \geq \cdots \geq \lambda_{r}$,
\beq\label{eq:generalized-eigenproblem}
H \psi_i = \lambda_i \Gamma \psi_i.
\eeq
Given $H$, which is practically computed in Section \ref{sec:practical-algorithms}, the generalized eigenvalue problem \eqref{eq:generalized-eigenproblem} can be efficiently solved by a randomized algorithm \cite{SaibabaLeeKitanidis2016} that only requires $O(r)$ matrix vector product.
We make the following key observation: \emph{The eigenvalue $\lambda_i$ measures the sensitivity of the data w.r.t.\ the parameters along direction $\psi_i$, i.e., the data mostly inform parameters in directions $\psi_i$ corresponding to large eigenvalues $\lambda_i$. For $i$ with small $\lambda_i$, close to zero, the variation of the likelihood $f$ in direction $\psi_i$ is negligible, so the posterior is close to the prior in direction $\psi_i$.}

We define a linear projector of rank $r$, $P_r: \bR^d \to \bR^d$, as 
\beq\label{eq:projector}
P_r x := \sum_{i=1}^r \psi_i \psi_i^T x = \Psi_r w, \quad \forall x \in \bR^d, 
\eeq
where $\Psi_r := (\psi_1, \dots, \psi_r) \in \bR^{d\times r}$ represents the projection matrix and $w := (w_1, \dots, w_r)^T \in \bR^r$ is the coefficient vector with element $w_i := \psi_i^T x$ for $i = 1, \dots, r$. For this projection, we seek a \emph{profile function} $g: \bR^d \to \bR$ such that $g(P_r x)$ is a good approximation of the likelihood function $f(x)$. For a given profile function, we define a projected density $p_r: \bR^d \to \bR$ as
\beq\label{eq:projected-posterior}
p_r(x) := \frac{1}{Z_r} g(P_r x) p_0(x),
\eeq
where $Z_r := \bE_{x \sim p_0}[g(P_r x)]$. It is shown in  \cite{ZahmCuiLawEtAl18} that an optimal profile function $g^*$ exists such that 
\beq
D_{\text{KL}} (p | p_r^{*}) \leq D_{\text{KL}} (p | p_r),
\eeq
where $p_r^*$ is defined as in \eqref{eq:projected-posterior} with an optimal profile function $g^*$. Moreover, under certain mild assumptions for the prior (sub-Gaussian) and the likelihood function (whose gradient has the second moment w.r.t.\ the prior), the upper bound is shown (sharper than that for pSVN in \cite{ChenWuChenEtAl19}) as
\beq\label{eq:projection-error-lambda}
D_{\text{KL}} (p | p_r^*) \leq \frac{\gamma}{2}\sum_{i = r+1}^d \lambda_i,
\eeq
for a constant $\gamma > 0$ independent of $r$, which implies small projection error for the posterior when $\lambda_i$ decay fast. The optimal profile function $g^*$ is nothing but the marginal likelihood given by 
\beq\label{eq:optimal-profile}
g^*(P_r x) = \int_{X_\perp} f(P_r x + \xi) p^\perp_0(\xi | P_r x) d \xi,
\eeq
 where $X_\perp$ is the complement of the subspace $X_r$ spanned by $\psi_1, \dots, \psi_r$, and
\beq\label{eq:marginals}
p^\perp_0(\xi | P_r x) = p_0(P_r x + \xi)/p^r_0(P_r x) \text{ with } p^r_0(P_r x) = \int_{X_\perp}  p_0( P_r x + \xi) d \xi.
\eeq
We defer a practical computation of the optimal profile function to Section \ref{sec:practical-algorithms}.

\subsection{Projected Stein Variational Gradient Descent}
\label{sec:w-pSVGD}

By the projection \eqref{eq:projector}, we consider a decomposition of the prior for the parameter $x= x^r + x^\perp$ as 
\beq
p_0(x) = p_0^r(x^r ) p_0^\perp(x^\perp|x^r),
\eeq
where the marginals $p_0^r$ and $p_0^\perp$ are defined in \eqref{eq:marginals}. Moreover, since $p_0^r$ only depends on $x^r = P_r x = \Psi_r w$, we define a prior density for $w$ as 
\beq\label{eq:w-prior}
\pi_0(w) = p_0^r(\Psi_r w).
\eeq
Then we define a joint (posterior) density for $w$ at the optimal profile function $g = g^*$ in \eqref{eq:projected-posterior} as 
\beq\label{eq:w-posterior}
\pi(w) = \frac{1}{Z_w} g(\Psi_r w) \pi_0(w),
\eeq
where the normalization constant $Z_w = E_{w \sim \pi_0}[g(\Psi_r w)]$. It is easy to see that $Z_w = Z_r $, where $Z_r $ is in \eqref{eq:projected-posterior}, and the projected density in \eqref{eq:projected-posterior} can be written as 
\beq\label{eq:w-Bayes}
p_r(x) = \pi(w) p_0^\perp(x^\perp|\Psi_r w).
\eeq
Therefore, to sample $x$ from $p_r$, we only need to sample $w$ from $\pi$ and $x^\perp$ from $p_0^\perp(x^\perp|\Psi_r w)$.

To sample from the posterior $\pi$ in \eqref{eq:w-posterior}, we employ the SVGD method presented in Section \ref{sec:SVGD} in the coefficient space $\bR^r$, with $r < d$. Specifically, we define a projected transport map $T^r: \bR^r \to \bR^r$ as 
\beq\label{eq:transport-w}
T^r(w) = w + \epsilon \phi^r(w),
\eeq
with a differentiable perturbation map $\phi^r: \bR^r \to \bR^r$, and a small enough $\epsilon > 0$ such that $T^r$ is invertible. Following the argument in \cite{LiuWang16} on the result \eqref{eq:directional-gradient} for SVGD, we obtain 
\beq
\nabla_\epsilon D_{\text{KL}}(T^r_\sharp \pi_0 | \pi) \big|_{\epsilon = 0} = - \bE_{w \sim \pi_0} [\text{trace}(\cA_{\pi} \phi^r(w))],
\eeq
where $\cA_{\pi}$ is the Stein operator for $\pi$ given by
\beq
\cA_{\pi} \phi^r (w)= \nabla_w \log \pi(w) \phi^r(w)^T + \nabla_w \phi^r(w).
\eeq

Using a tensor product of RKHS $\cH_r$ with kernel $k^r(\cdot, \cdot): \bR^r \times \bR^r \to \bR$ for the approximation of $\phi^r \in (\cH_r)^r = \cH_r \times \cdots \times \cH_r$,  a SVGD update of the samples $w_1^0, \dots, w_N^0$ from $\pi_0(w)$ leads to
\beq\label{eq:w-update-l}
w_m^{\ell+1} = w_m^\ell + \epsilon_l \hat{\phi}_{\ell}^{r, *}(w_m^\ell), \quad m = 1, \dots, N, \ell = 0, 1, \dots,
\eeq
with a step size $\epsilon_l$ and an approximate steepest direction 
\beq\label{eq:w-gradient-saa}
\hat{\phi}_{\ell}^{r, *}(w_m^\ell) = \frac{1}{N} \sum_{n=1}^N \nabla_{w_n^\ell} \log \pi(w_n^\ell) k^r(w_n^\ell, w_m^\ell) + \nabla_{w_n^\ell} k^r(w_n^\ell, w_m^\ell).
\eeq

The kernel $k^r$ can be specified as in \eqref{eq:x-kernel}, i.e.,
\beq\label{eq:w-kernel}
k^r(w,w') = \exp\left(- \frac{||w-w'||^2_2}{h}\right).
\eeq
To account for data impact in different directions $\psi_1, \dots, \psi_r$ informed by the eigenvalues of \eqref{eq:generalized-eigenproblem}, we propose to replace $||w-w'||^2_2$ in \eqref{eq:w-kernel} by $(w-w')^T (\Lambda + I) (w - w')$ with $\Lambda = \text{diag}(\lambda_1, \dots, \lambda_r)$ for the likelihood and $I$ for the prior.

The following theorem, proved in Appendix \ref{sec:pSVGD-proof}, gives $\nabla_w \log \pi(w)$ and the connection between pSVGD for the coefficient $w$ and SVGD for the projected parameter $P_r x$ under certain conditions.
\begin{theorem}
	\label{thm:pSVGD}
	The gradient of the posterior $\pi$ in \eqref{eq:w-posterior} is given by 
	\beq\label{eq:w-grad-post}
	\nabla_w \log \pi(w) = \Psi_r^T \left(\frac{\nabla_x g(P_r x) }{g(P_r x)} + \frac{\nabla_x p_0^r(P_r x)}{p_0^r(P_r x)} \right).
	\eeq
	Moreover,  with the kernel $k^r(\cdot, \cdot): \bR^r \times \bR^r \to \bR$ defined in \eqref{eq:w-kernel} and $k(\cdot, \cdot): \bR^d \times \bR^d \to \bR$ defined in \eqref{eq:x-kernel}, if $p_0^r(P_r x) = p_0(P_r x)$, for example $p_0$ is Gaussian, we have the equivalence of the projected transport map $T^r$ for the coefficient $w$
	and the transport map $T$ for the projected parameter $P_r x$, as
	\beq\label{eq:projection-equivalence}
	T^r(w) = \Psi_r^T T(P_r x).
	\eeq
	In particular, we have 
	\beq\label{eq:log-posterior-equivalence}
	\nabla_w \log \pi(w) = \Psi_r^T \nabla_x \log p_r(P_r x).
	\eeq
\end{theorem}

\subsection{Practical algorithms}
\label{sec:practical-algorithms}

Sampling from the projected density $p^*_r(x)$ defined in \eqref{eq:projected-posterior} for the optimal profile function $g^*$ in \eqref{eq:optimal-profile} involves, by the decomposition \eqref{eq:w-Bayes}, sampling $w$ from the posterior $\pi$ by pSVGD and sampling $x^\perp$ from the conditional distribution with density $p^\perp_0(x^\perp | \Psi_r w)$. The sampling is impractical because of two challenges:
	(1) Both $p^\perp_0(x^\perp | \Psi_r w)$ and $g^*$ in \eqref{eq:optimal-profile} involve high-dimensional integrals.
	(2) The matrix $H$ defined in \eqref{eq:grad-grad-T} for the construction of the basis $\psi_1, \dots, \psi_r$ involves integration w.r.t.\ the posterior distribution of the parameter $x$. However, drawing samples from the posterior to evaluate the integral turns out to be the central task of the Bayesian inference.

The first challenge can be practically addressed by using the property --- the posterior distribution only effectively differs from the prior in the dominant subspace $X_r$, or the variation of likelihood $f$ in the complement subspace $X_\perp$ is negligible. 
%
Therefore, for any sample $x_n^0$ drawn from the prior $p_0$, $n = 1, \dots, N$, we compute $x_n^\perp = x_n^0 - P_r x_n^0$ and freeze it for given $P_r$ as a sample from $p^\perp_0(x^\perp | P_r x)$. Moreover, at sample $x_n^\ell$ we approximate the optimal profile function $g^*$ in \eqref{eq:optimal-profile} as
\beq
g^*(P_r x_n^\ell) \approx f(P_r x_n^\ell + x_n^\perp),
\eeq
which is equivalent to using one sample $x_n^\perp$ to approximate the integral \eqref{eq:optimal-profile} because the variation of $f$ in the complement subspace $X_\perp$ is small. This is used in computing $\nabla_w\log \pi(w)$ in \eqref{eq:w-grad-post}.

Given the projector $P_r$ with basis $\Psi_r$, we summarize the pSVGD transport of samples in Algorithm \ref{alg:pSVGD}. 
In particular, by leveraging the property that the samples can be updated in parallel, we implement a parallel version of pSVGD using MPI for information communication in $K$ processor cores, each with $N$ different samples, thus producing $M = N K$ different samples in total. 

\begin{algorithm}[!htb]
	\caption{pSVGD in parallel}
	\label{alg:pSVGD}
	\begin{algorithmic}[1]
		\STATE{\bfseries Input:} samples $\{x_n^0\}_{n=1}^N$ in each of $K$ cores, basis $\Psi_r$, maximum iteration $L_{\text{max}}$, tolerance $w_{\text{tol}}$.
		\STATE{\bfseries Output:} posterior samples $\{x_n^*\}_{n=1}^N$ in each core.
		\STATE Set $\ell = 0$, project $w_n^0 = \Psi_r^T x_n^0$,  $x_n^\perp = x_n^0 - \Psi_r w_n^0$, and perform {MPI\_Allgather} for $\{w_n^0\}_{n=1}^N$.
		\REPEAT
		\STATE Compute gradients $\nabla_{w_n^\ell} \log \pi(w_n^\ell)$ by \eqref{eq:w-grad-post} for $n = 1, \dots, N$, and perform {MPI\_Allgather}.
		\STATE Compute the kernel values $k^r(w_n^\ell, w_m^\ell)$ and their gradients $\nabla_{w_n^\ell} k^r(w_n^\ell, w_m^\ell)$ for $n = 1, \dots, NK$, $m = 1, \dots, N$, and perform {MPI\_Allgather} for them.
		\STATE Update samples $w^{\ell+1}_m$ from $w^{\ell}_m$ by \eqref{eq:w-update-l} and \eqref{eq:w-gradient-saa} for $m = 1, \dots, N$, with $NK$ samples used for SAA in \eqref{eq:w-gradient-saa}, and perform {MPI\_Allgather} for $\{w_m^0\}_{m=1}^N$.
		\STATE Set $\ell \leftarrow \ell + 1$.
		\UNTIL{$\ell \geq L_{\text{max}}$ or $\text{mean} (||w_m^\ell - w_m^{\ell-1}||_2) \leq w_{\text{tol}}$.}
		\STATE Reconstruct samples $x_n^* = \Psi_r w_n^\ell + x_n^\perp$.
	\end{algorithmic}
\end{algorithm}

To construct the projector $P_r$ with basis $\Psi_r$, we approximate $H$ in \eqref{eq:grad-grad-T} by 
\beq\label{eq:a-H}
\hat{H}: = \frac{1}{M}\sum_{m=1}^M \nabla_x \log f(x_m) (\nabla_x \log f(x_m))^T.
\eeq
where $x_1, \dots, x_M$ are supposed to be samples from the posterior, which are however not available at the beginning. We propose to adaptively construct the basis $\Psi_r^\ell$ with samples $x_1^\ell, \dots, x_M^\ell$ transported from the prior samples $x_1^0, \dots, x_M^0$ by pSVGD. This procedure is summarized in Algorithm \ref{alg:apSVGD}. We remark that by the adaptive construction, we push the samples to their posterior in each subspace $X_r^{\ell_x}$ spanned by (possibly) different basis $\Psi_r^{\ell_x}$ with different $r$ for different ${\ell_x}$, during which the frozen samples $x^\perp_n$ in Algorithm \ref{alg:pSVGD} are also updated at each step $\ell_x$ of Algorithm \ref{alg:apSVGD}. 

\begin{algorithm}[!htb]
	\caption{Adaptive pSVGD in parallel}
	\label{alg:apSVGD}
	\begin{algorithmic}[1]
		\STATE{\bfseries Input:} samples $\{x_n^0\}_{n=1}^N$ in each of $K$ cores, $L_{\text{max}}^x, L^w_{\text{max}}$, $x_{\text{tol}}, w_{\text{tol}}$.
		\STATE{\bfseries Output:} posterior samples $\{x_n^*\}_{n=1}^N$ in each core.
		\STATE Set $\ell_x = 0$.
		\REPEAT
		\STATE Compute $\nabla_x \log f(x_n^{\ell_x})$ in \eqref{eq:a-H} for $n = 1, \dots, N$ in each core, and perform {MPI\_Allgather}. 
		\STATE Solve \eqref{eq:generalized-eigenproblem} with $H$ approximated as in \eqref{eq:a-H}, with all $M = NK$ samples, to get bases $\Psi_r^{\ell_x}$.
		\STATE Apply the pSVGD Algorithm \ref{alg:pSVGD}, i.e., 
		$$
		\{x_n^*\}_{n=1}^N = \text{pSVGD}(\{x_n^{\ell_x}\}_{n=1}^N, \Psi_r^{\ell_x}, L_{\text{max}}^w, w_{\text{tol}}).
		$$
		\STATE Set $\ell_x \leftarrow \ell_x + 1$ and $x_n^{\ell_x} = x_n^*$, $n = 1, \dots, N$.
		\UNTIL{$\ell_x \geq L^x_{\text{max}}$ or $\text{mean} (||x_m^{\ell_x} - x_m^{\ell_x-1}||_X) \leq x_{\text{tol}}$.}
	\end{algorithmic}
\end{algorithm}


\section{Numerical Experiments}
\label{sec:numerics}

We present three Bayesian inference problems with high-dimensional parameters to demonstrate the accuracy of pSVGD compared to SVGD, and the convergence and scalability of pSVGD w.r.t.\ the number of parameters, samples, data points, and processor cores. A linear inference example, whose posterior is analytically given, is presented in Appendix \ref{sec:linear-inference} to demonstrate the accuracy of pSVGD. An application in COVID-19 to infer the social distancing effect given hospitalized data is presented in Appendix \ref{sec:covid19-inference}. The code is available at {\url{https://github.com/cpempire/pSVGD}}.

\subsection{Conditional diffusion process}
\begin{figure}[!htb]
	\vskip -0.1in
	\centering
	\makebox[\linewidth][c]{
		\includegraphics[width=0.55\linewidth]{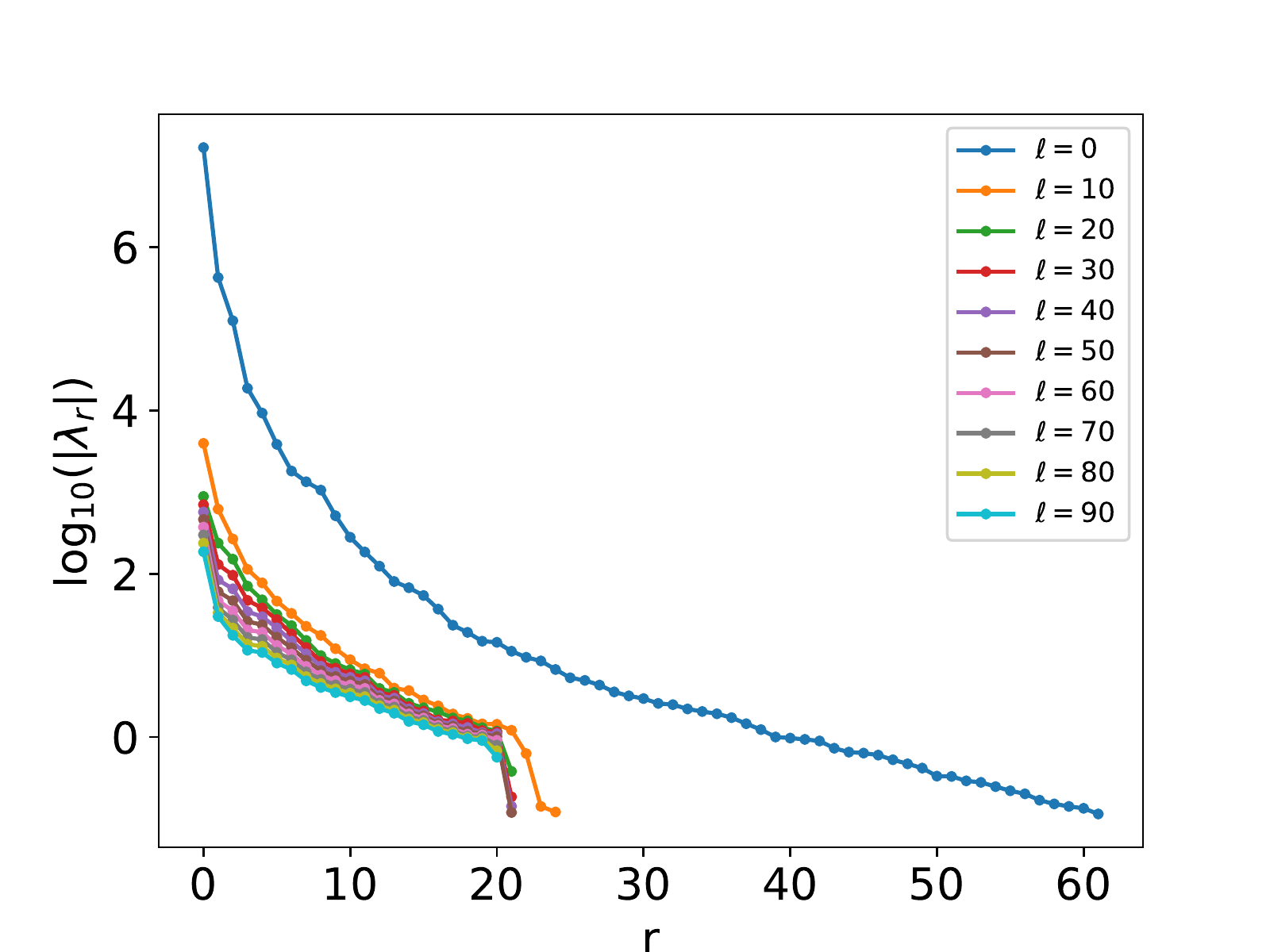}\hskip -0.3in
		\includegraphics[width=0.55\linewidth]{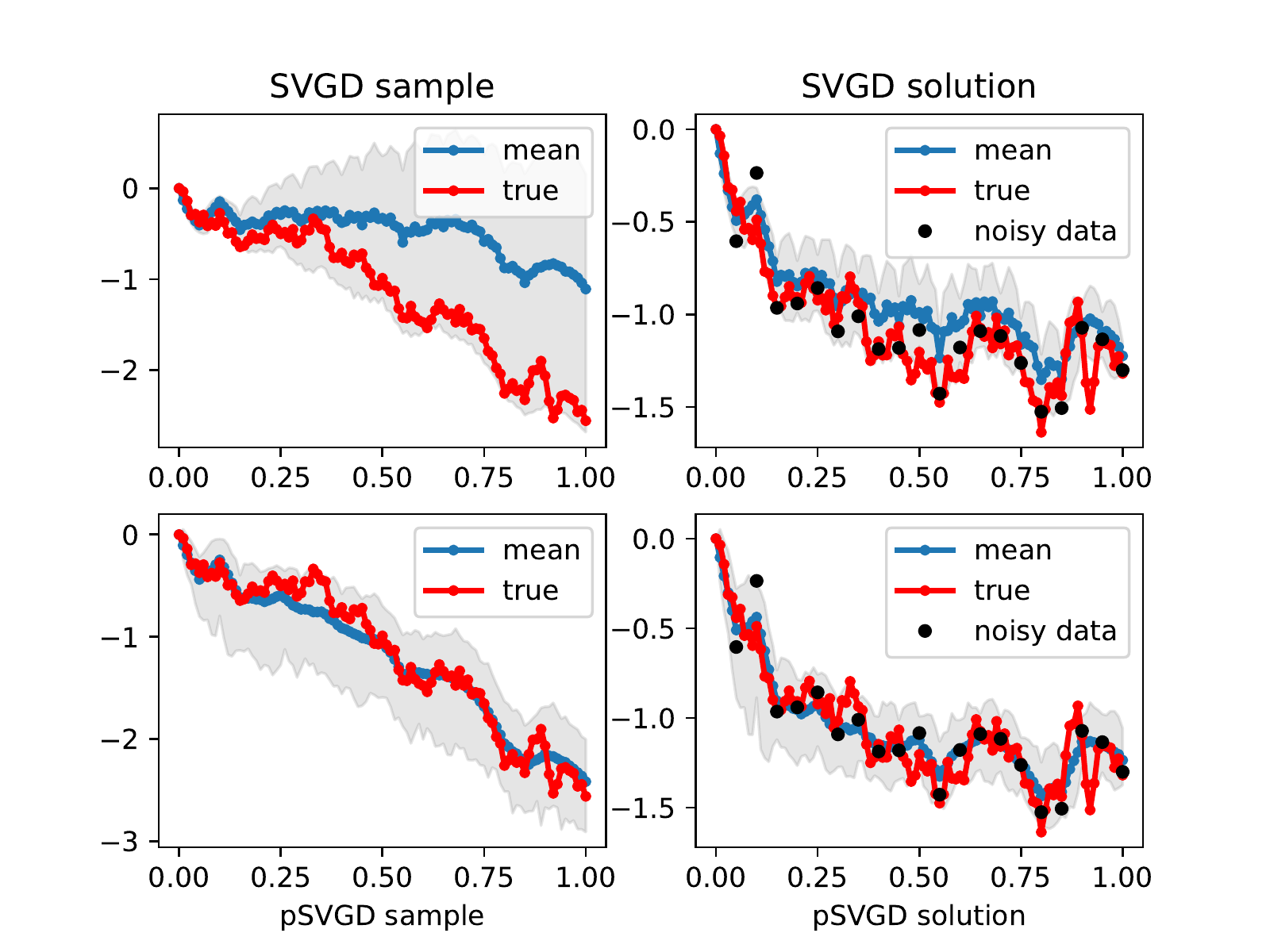}
	}
    \vskip -0.1in
	\caption{Left: Decay of the eigenvalues of \eqref{eq:generalized-eigenproblem} at different iteration numbers $\ell$. Right: SVGD (top) and pSVGD (bottom) samples and solutions at iteration $\ell = 100$, including the synthetic true, posterior mean, 90\% confidence interval in shadow, and noisy data points. }\label{fig:conditional-diffusion}
	\vskip -0.2in
\end{figure}

We consider a high-dimensional model that is often used to test inference algorithms in high dimensions, e.g., Stein variational Newton in \cite{DetommasoCuiMarzoukEtAl18}, which is discretized from a conditional diffusion process
\beq\label{eq:conditional-diffusion}
du_t = \frac{10 u (1-u^2)}{1+u^2} dt + dx_t, \quad t \in (0, 1],
\eeq
with zero initial condition $u_0 = 0$.
The forcing term $(x_t)_{t\geq 0}$ is a Brownian motion, whose prior is Gaussian with zero mean and covariance $C(t, t') = \min(t, t')$. We use Euler-Maruyama scheme with step size $\Delta t  = 0.01$ for the discretization, which leads to dimension $d = 100$ for the discrete Brownian path $x$. We generate the data by first solving \eqref{eq:conditional-diffusion} at a true Brownian path $x_{\text{true}}$, and taking $y =(y_1, \dots, y_{20})$ with $y_i = u_{t_{i}} + \xi_i$ for equispaced $t_1, \dots, t_{20}$ in $(0, 1]$ and additive noise $\xi_i \in N(0, \sigma^2)$ with $\sigma = 0.1$. We run SVGD and the adaptive pSVGD with line search for the learning rate, using $N =128$ samples to infer $x_{\text{true}}$, where the subspace for pSVGD is updated every 10 iterations. The results are displayed in Figure \ref{fig:conditional-diffusion}. From the left we can see a fast decay of eigenvalues of \eqref{eq:generalized-eigenproblem}, especially when the iteration number $\ell$ becomes big with the samples converging to the posterior, which indicates the existence of an intrinsic low-dimensional subspace. From the right we can observe that pSVGD leads to samples at which the solutions are much closer to the noisy data as well as the true solution than that of SVGD. Moreover, the posterior mean of pSVGD is much closer to the true parameter with much tighter 90\% confidence interval covering $x_{\text{true}}$ than that of SVGD.

\subsection{Bayesian logistic regression}
\begin{figure}[!htb]
	\vskip -0.1in
	\centering
	\makebox[\linewidth][c]{
		\includegraphics[width=0.55\linewidth]{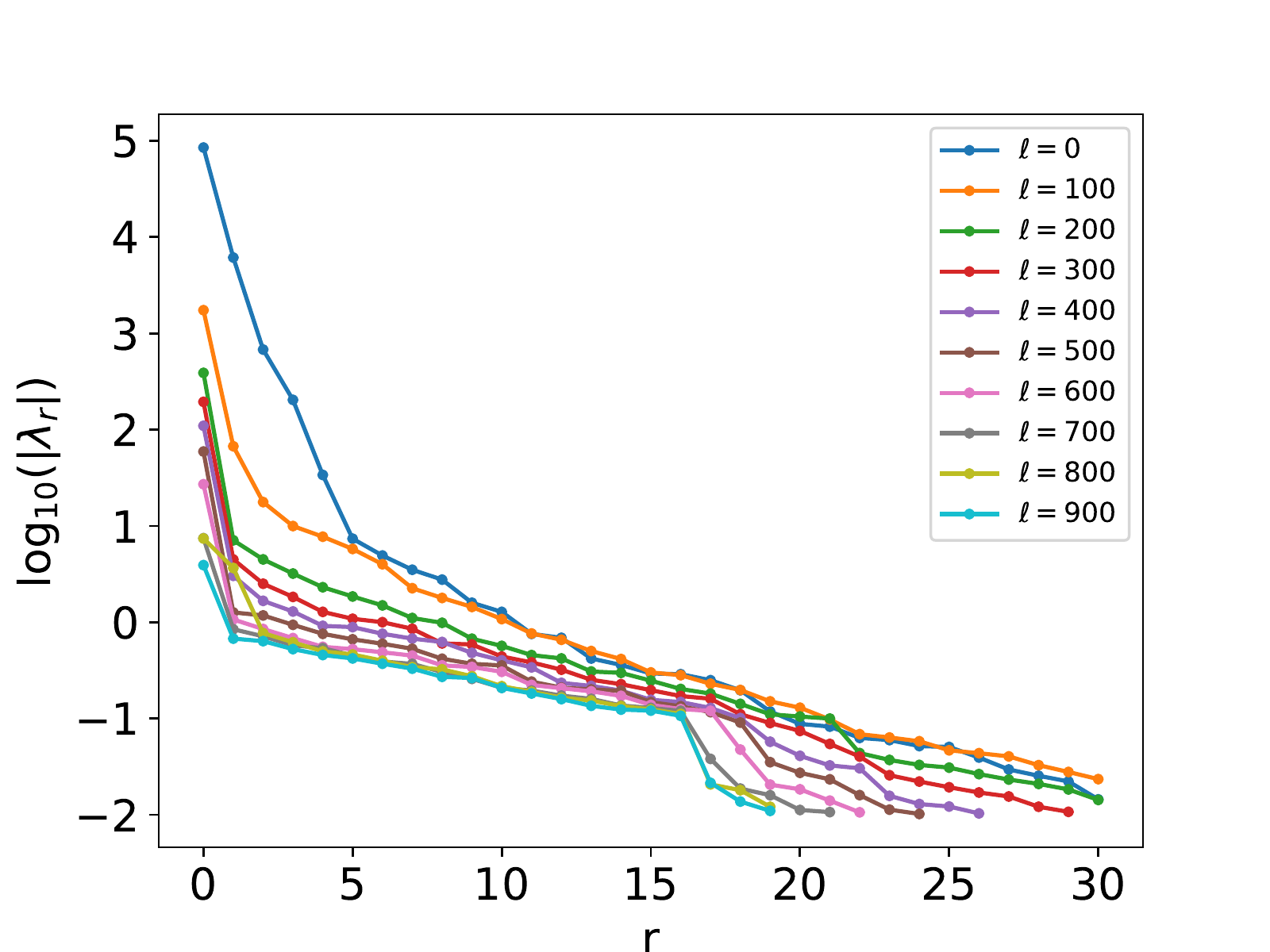}\hskip -0.3in
		\includegraphics[width=0.55\linewidth]{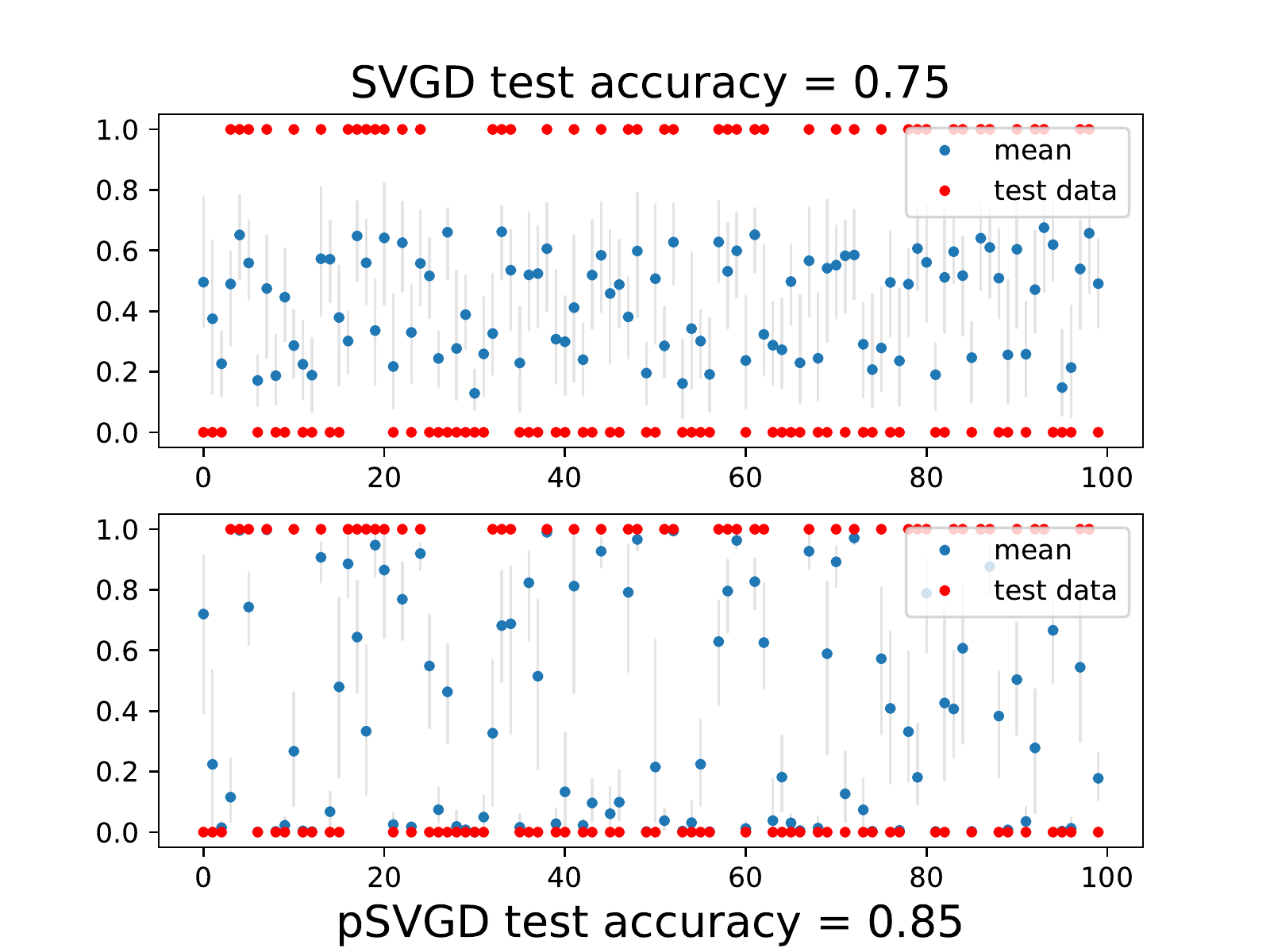}
	}
	\vskip -0.1in
	\caption{Left: Decay of the eigenvalues of \eqref{eq:generalized-eigenproblem} at different iteration numbers $\ell$. Right: SVGD (top) and pSVGD (bottom) test data and prediction at iteration $\ell = 1000$, including posterior mean, 90\% confidence interval in shadow. The 85\% test accuracy for pSVGD is the same as for SVM from the data source file. The training time for pSVGD is 201 seconds compared to 477 seconds for SVGD. }\label{fig:logistic-regression}
	\vskip -0.1in
\end{figure}

We consider Bayesian logistic regression for  binary classification of cancer and normal patterns for mass-spectrometric data with $10,000$ attributes from  \href{https://archive.ics.uci.edu/ml/datasets/Arcene}{https://archive.ics.uci.edu/ml/datasets/Arcene}, which leads to $d = 10,000$ parameters (with i.i.d.\ uniform distribution as prior for Figure \ref{fig:logistic-regression}; Gaussian is also tested with similar results). We use 100 data for training and 100 for testing. We run pSVGD and SVGD with line search and 32 samples, with projection basis updated every 100 iterations. The results are shown in Figure \ref{fig:logistic-regression}. We can see a dramatic decay of the eigenvalues, which indicates an intrinsic dominant low dimensional subspace in which pSVGD effectively drives the samples to the posterior, and leads to more accurate prediction than that of SVGD. We remark that 32 samples in the estimate for the gradient information matrix in \eqref{eq:a-H} are sufficient to capture the subspace since more samples lead to similar decay of eigenvalues as in Figure \ref{fig:logistic-regression}.
\subsection{Partial differential equation}
In this example we consider an elliptic partial differential equation model (widely used in various fields, e.g., inference for permeability in groundwater flow, thermal conductivity in material science, electrical impedance in medical imaging, etc.), with a simplified form as 
\beq\label{eq:nonlinear-pde}
-\nabla \cdot (e^{\mathrm{x}} \nabla \mathrm{u}) = 0, \quad \text{in } (0, 1)^2,
\eeq
which is imposed with Dirichlet boundary conditions $\mathrm{u} = 1$ on the top boundary and $\mathrm{u} = 0$ on bottom boundary, and homogeneous Neumann boundary conditions on the left and right boundaries.
$\nabla \cdot$ is a divergence operator, and $\nabla$ is a gradient operator. $\mathrm{x}$ and $\mathrm{u}$ are discretized by finite elements with piecewise linear elements in a uniform mesh of triangles of size $d$. $x\in \bR^d$ and $u \in \bR^d$ are the nodal values of $\mathrm{x}$ and $\mathrm{u}$. We consider a Gaussian distribution for $\mathrm{x} \in \cN(0, \cC)$ with covariance $\cC = (-0.1\Delta + I)^{-2}$, which leads to a Gaussian distribution for $x \sim \cN(0, \Sigma_x)$, where $\Sigma_x \in \bR^{d\times d}$ is discretized from $\cC$. We consider a parameter-to-observable map 
$h(x) = O \circ S(x),$
where $S: x \to u$ is a nonlinear discrete solution map of the equation \eqref{eq:nonlinear-pde}, $O:\bR^d \to \bR^s$ is a pointwise observation map at $s = 7\times 7$ points equally distributed in $(0, 1)^2$. We consider an additive $5\%$ noise $\xi \sim \cN(0, \Sigma_\xi)$ with $\Sigma_\xi = \sigma^2 I$ and $\sigma = \max (|O u|)/20$ for data 
$
y = h(x) + \xi.
$

\begin{figure}[!htb]
	\vskip -0.2in	
	\centering
	\makebox[\linewidth][c]{
	\includegraphics[width=0.32\textwidth]{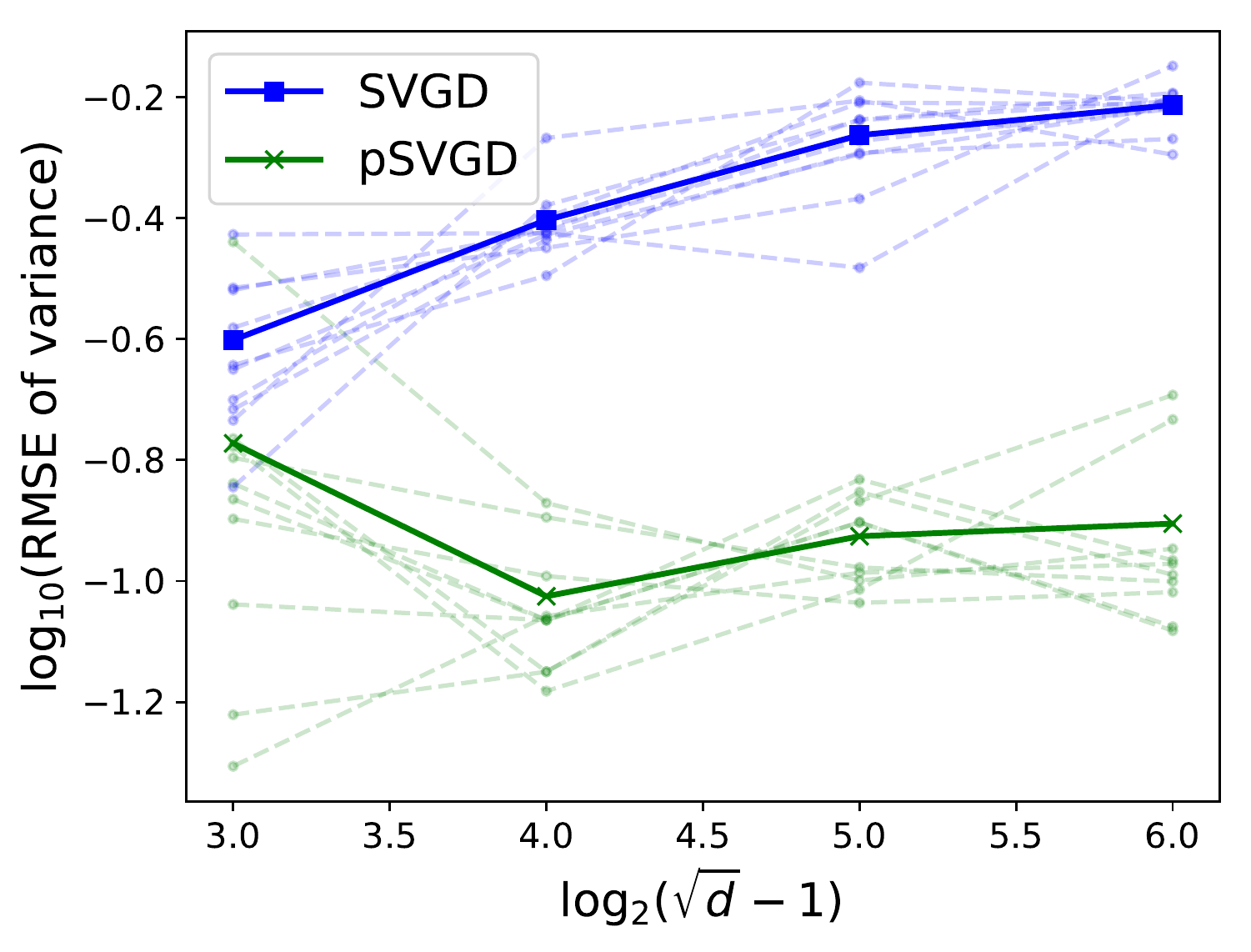} 
		\includegraphics[width=0.37\textwidth]{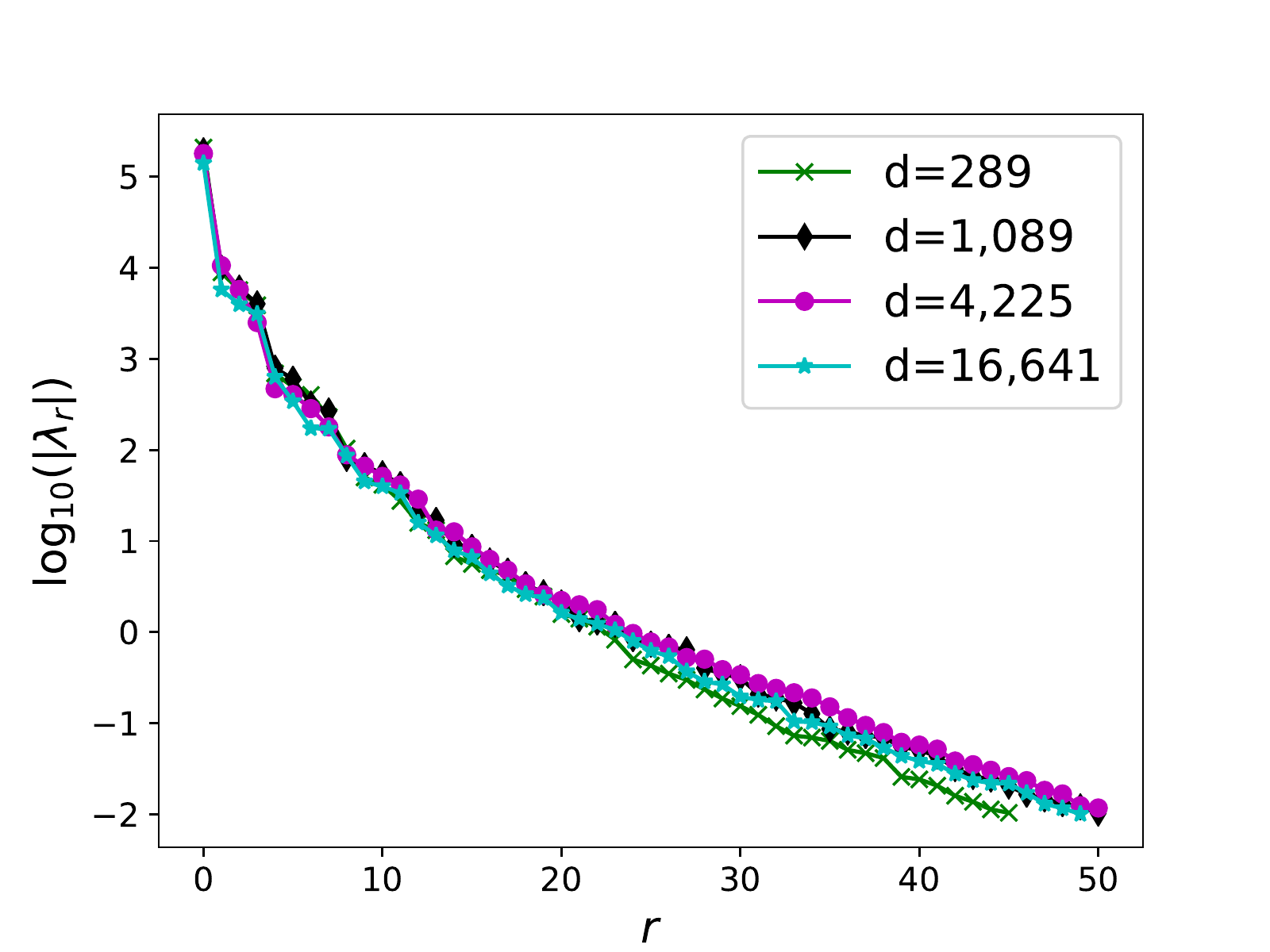} \hskip -0.2in
	\includegraphics[width=0.37\textwidth]{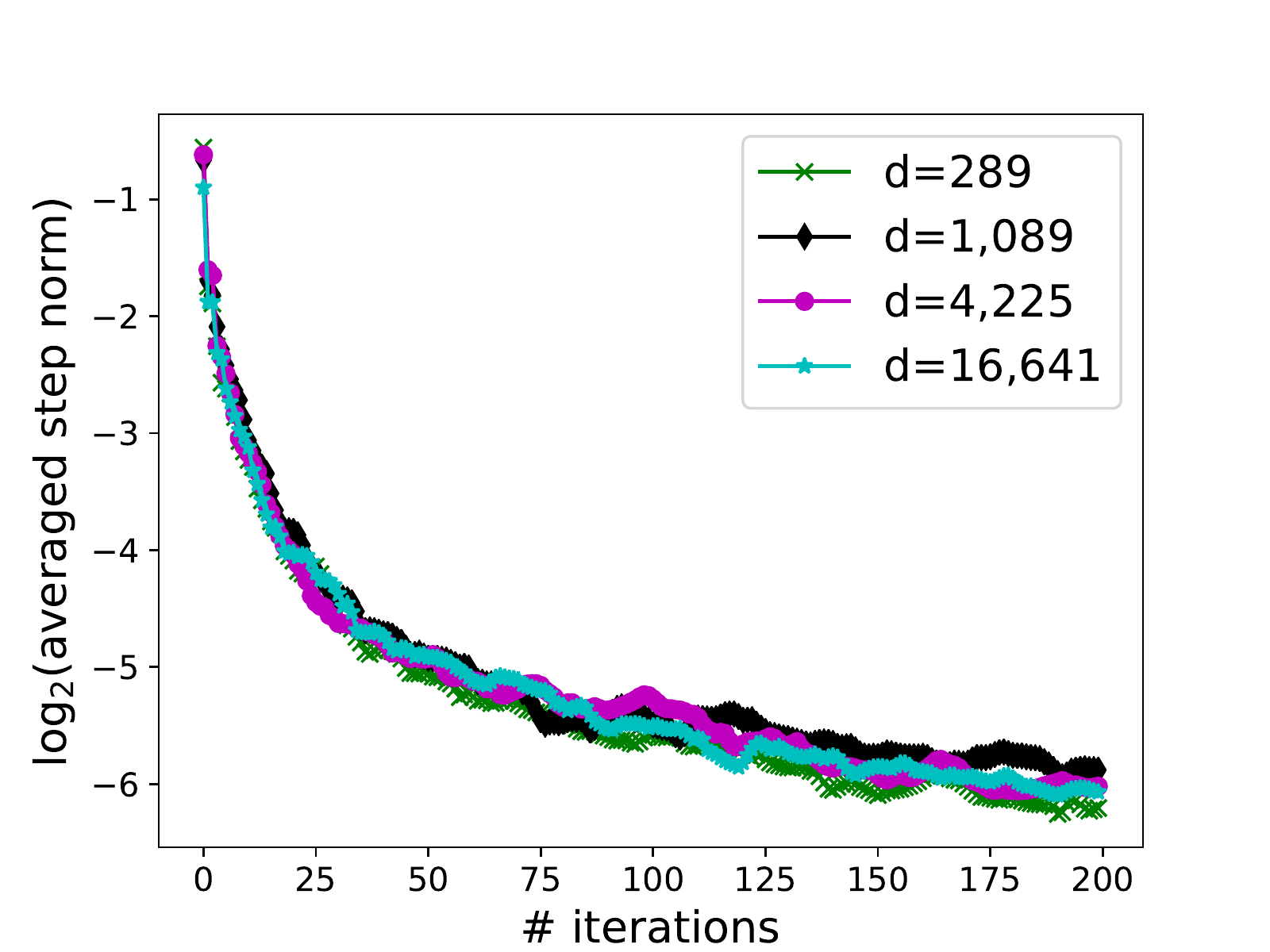}
	}
	\vskip -0.1in
	\caption{Left: RMSE of pointwise sample variance in $L_2$-norm, with dimension $d = (2^n+1)^2$, $n = 3, 4, 5, 6$. Middle: Scalability w.r.t.\ $d$ by decay of eigenvalues $\lambda_r$ w.r.t.\ $r$. Right: decay of the averaged step norm $\text{mean}_m||w^{\ell+1}_m-w^\ell_m||_2$ w.r.t.\ the number of iterations for different dimension $d$.}
	\label{fig:accuracy-nonlinear}
	\vskip -0.2in	
\end{figure}
We use a DILI-MCMC algorithm \cite{CuiLawMarzouk16} to generate $10,000$ effective posterior samples and use them to compute a reference sample variance. 
We run SVGD and the adaptive pSVGD (with $\lambda_{r+1} < 10^{-2}$) using 256 samples and 200 iterations for different dimensions,  both using line search to seek the step size $\epsilon_\ell$. The comparison of accuracy can be observed in the left of Figure \ref{fig:accuracy-nonlinear}. We can see that SVGD samples fail to capture the posterior distribution in high dimensions and become worse with increasing dimension, while pSVGD samples represent the posterior distribution well, measured by sample variance, and the approximation remains accurate with increasing dimension.
\begin{figure}[!htb]
	\centering
	\vskip -.15in		\makebox[\linewidth][c]{\includegraphics[width=0.37\textwidth]{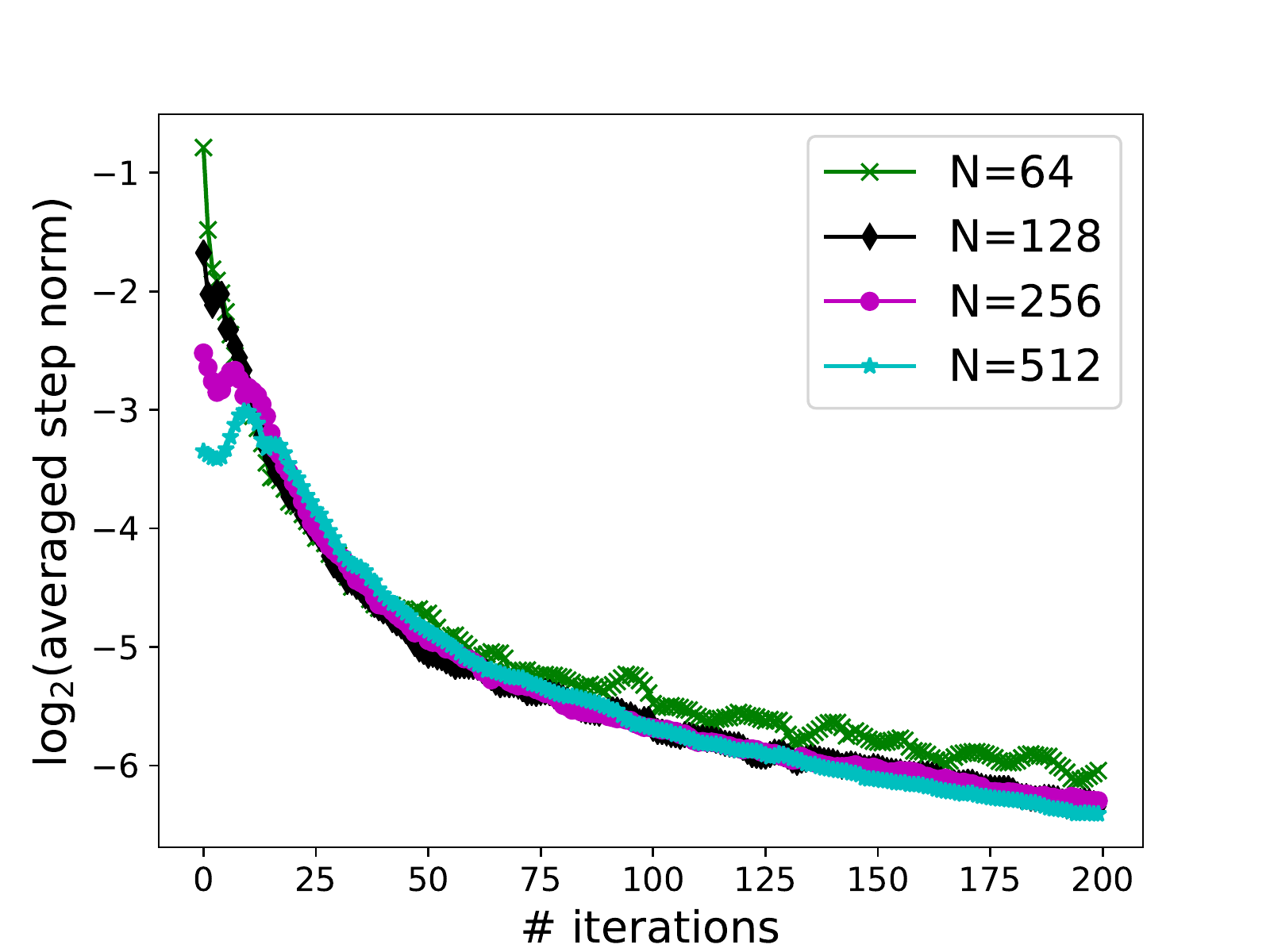}\hskip -0.2in
		\includegraphics[width=0.37\textwidth]{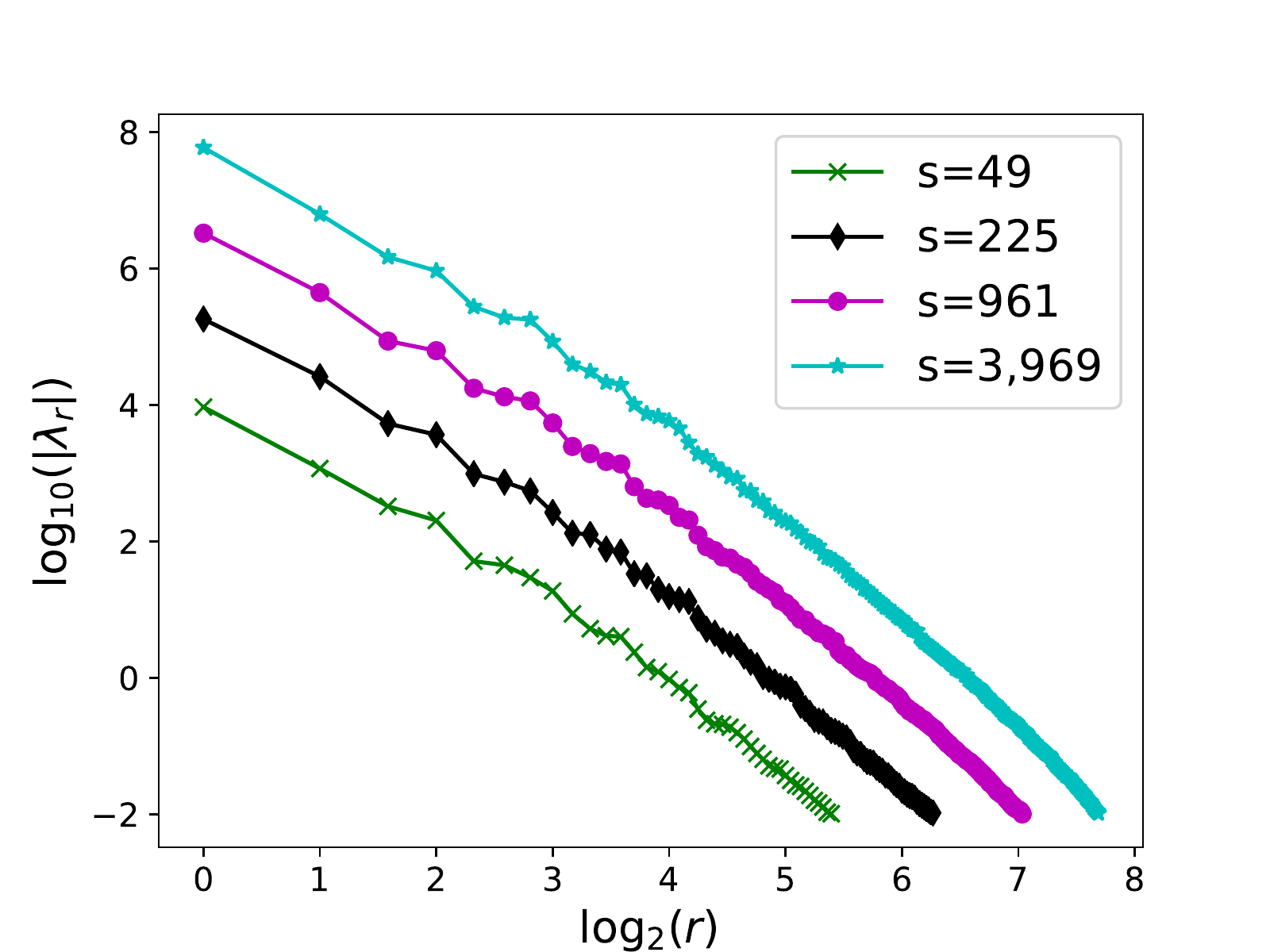}\hskip -0.2in
		\includegraphics[width=0.37\textwidth]{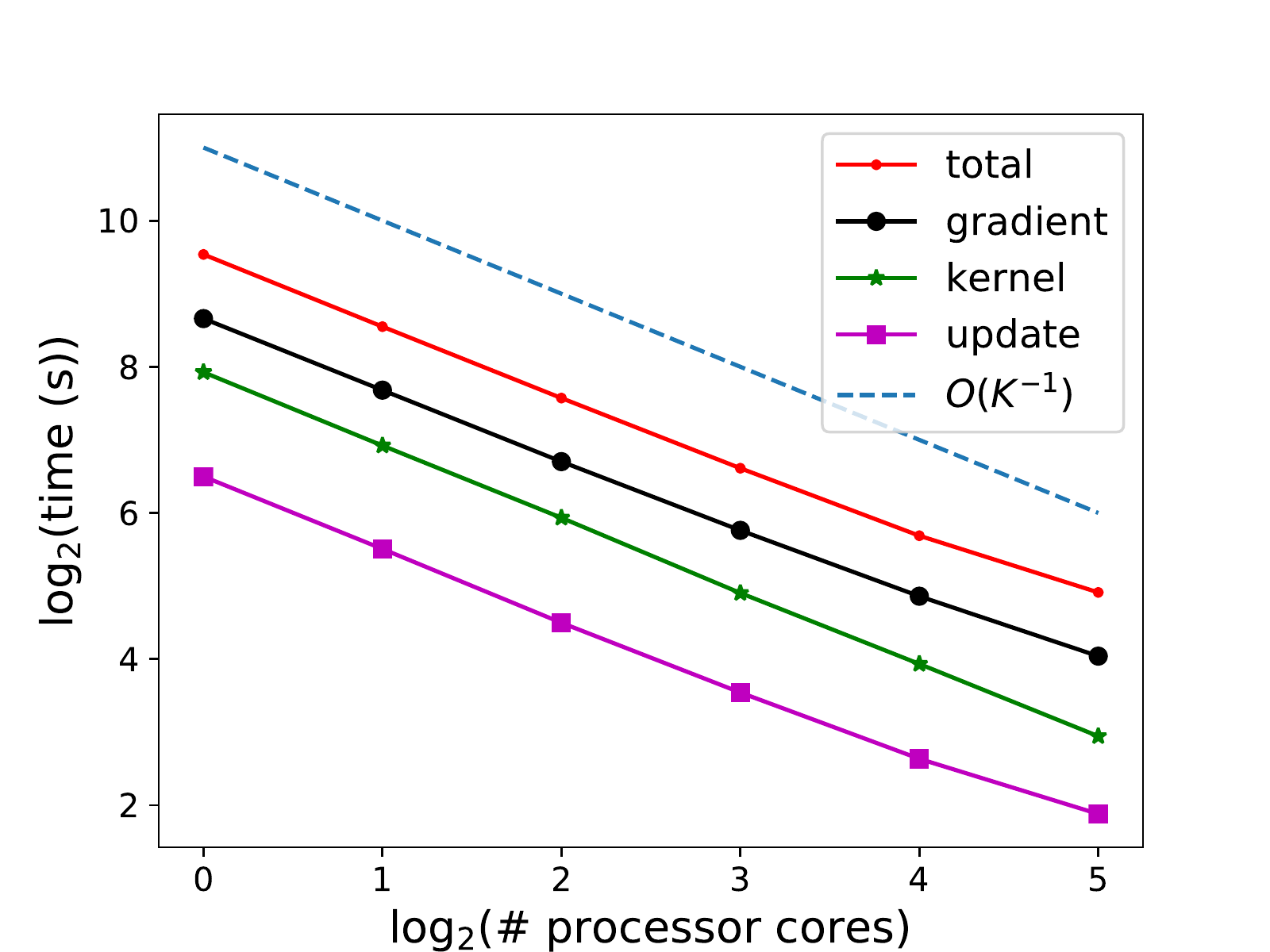}}
	\vskip -0.1in	
	\caption{Scalability w.r.t.\ the number of (1) samples $N$ by decay of the averaged step norm (left), (2) data points $s$ by decay of eigenvalues (middle), and (3) processor cores $K$ by decay of CPU time (for gradient including eigendecompostion \eqref{eq:generalized-eigenproblem}, kernel, sample update, total) of pSVGD (right).
	}
	\label{fig:scalability-data-core}
	\vskip -0.1in	
\end{figure}
The accuracy of pSVGD can be further demonstrated by the significant decay (about 7 orders of magnitude) of the eigenvalues for different dimensions in the middle of Figure \ref{fig:accuracy-nonlinear}. Only about 50 dimensions (with small relative projection error, about $\cE_r < 10^{-6}$, committed in the posterior by \eqref{eq:projection-error-lambda}) are preserved out of from 289 to 16,641 dimensions, representing over 300$\times$ dimension reduction for the last case. The similar decays of the eigenvalues $\lambda_r$ in the projection rank $r$ and the averaged step norm $\text{mean}_m||w^{\ell+1}_m-w^\ell_m||_2$ in the number of iterations shown in the right of Figure \ref{fig:accuracy-nonlinear} imply that pSVGD is scalable w.r.t.\ the parameter dimension. Moreover, the similar decays for different sample size $N = 64, 128, 256, 512$ in the left of Figure \ref{fig:scalability-data-core} demonstrate that pSVGD is scalable w.r.t.\ the number of samples $N$. Furthermore, as displayed in the middle of Figure \ref{fig:scalability-data-core}, with increasing number of i.i.d.\ observation data points $s = 7^2, 15^2, 31^2, 63^2$ in a refined mesh of size $d = 17^2, 33^2, 65^2, 129^2$, the eigenvalues decay at almost the same rate with similar relative projection error $\cE_r$, and lead to similar reduction $d/r$ for $r$ such that $\lambda_{r+1}< 10^{-2}$, which implies weak scalability of pSVGD w.r.t.\ the number of data points. Lastly, from the right of Figure \ref{fig:scalability-data-core} by the nearly $O(K^{-1})$ decay of CPU time we can see that pSVGD achieves strong parallel scalability (in computing gradient, kernel, and sample update) w.r.t.\ the number of processor cores $K$ for the same work with $KN = 1024$ samples.

\section{Conclusions}
We proposed a new algorithm --- pSVGD for Bayesian inference in high dimensions to tackle the critical challenge of the curse of dimensionality. The projection error committed in the posterior can be bounded by the truncated (fast decaying) eigenvalues. We proved that pSVGD for the coefficient is equivalent to SVGD for the projected parameter under suitable assumptions. 
We demonstrated that pSVGD overcomes the curse of dimensionality for several high-dimensional Bayesian inference problems. In particular, we showed that pSVGD is scalable w.r.t.\ the number of parameters, samples, data points, and processor cores for a widely used benchmark problem in various scientific and engineering fields, which is crucial for solving high-dimensional and large-scale inference problems. 


\bibliographystyle{plain}
\bibliography{bibliographyAuth3Year}



\appendix



\section{Proof of Theorem \ref{thm:pSVGD}}
\label{sec:pSVGD-proof}
\begin{proof}
	We first show \eqref{eq:w-grad-post}. By definition of $\pi(w)$ in \eqref{eq:w-posterior}, we have 
	\beq\label{eq:split-w}
	\nabla_w \log \pi(w) = \nabla_w \log(g(\Psi_r w)) + \nabla_w \log(\pi_0(w)).
	\eeq
	For the second term, by definition of $\pi_0$ in \eqref{eq:w-prior}, we have 
	\beq
	\nabla_w \log(\pi_0(w)) = \nabla_w \log p_0^r(\Psi_r w) = \frac{\nabla_w p_0^r(\Psi_r w)}{p_0^r(\Psi_r w)},
	\eeq
	where by definition of $p_0^r(\Psi_r w)$ in \eqref{eq:marginals} we have 
	\beq
	\nabla_w p_0^r(\Psi_r w) =  \int_{X_\perp} \nabla_w p_0(\Psi_r w + \xi) d\xi =  \int_{X_\perp} \Psi_r^T \nabla_x p_0(\Psi_r w + \xi) d\xi = \Psi_r^T \nabla_x p_0^r(\Psi_r w). 
	\eeq 
	The first term $\nabla_w \log(g(\Psi_r w))$ in \eqref{eq:split-w} can be derived similarly by the definition of $g$ in \eqref{eq:optimal-profile} and \eqref{eq:marginals}. 
	
	Next, we proceed to prove the equivalence \eqref{eq:projection-equivalence} and \eqref{eq:log-posterior-equivalence} at the first step $\ell = 0$. Then the equivalence for steps $\ell > 0$ follows by induction. 
	By the parameter decomposition $x = x^r + x^\perp$ with $x^r = P_r x$, we denote $\eta_0$ and $\eta$ as the prior and posterior for the projected parameter $x^r$, given by
	\beq\label{eq:pi-p}
	\eta_0(x^r) = p_0(P_r x) \text{ and } \eta(x^r) = p_r(P_r x), 
	\eeq
	where $p_r$ is the projected density defined in \eqref{eq:projected-posterior} with optimal profile function $g = g$ given in \eqref{eq:optimal-profile}. Equivalently, by the property of the projection $P_r P_r x = P_r x $, we have 
	\beq
	\eta(x^r) = \frac{1}{Z_r} g(x^r) \eta_0(x^r).
	\eeq
	
	We can write the transport map \eqref{eq:transport} for the projected parameter $x^r$ in the steepest direction $ \varphi_{0}$ as
	\beq
	T(x^r) = x^r + \epsilon \varphi_{0}(x^r),
	\eeq
	where $\varphi_{0}$ is given by 
	\beq
	\varphi_{0}(\cdot) = \bE_{x^r \sim \eta_0}[\cA_{\eta} \kappa(x^r, \cdot)],
	\eeq
	with the kernel $\kappa(x^r, \Tilde{x}^r) = k(P_r x, P_r \Tilde{x})$ for any $x, \Tilde{x} \in \bR^d$ and the Stein operator 
	\beq
	\cA_\eta \kappa(x^r, \cdot) = \nabla_{x^r} \log \eta(x^r) \kappa(x^r, \cdot) + \nabla_{x^r} \kappa(x^r, \cdot).
	\eeq
	By definition of the kernel in \eqref{eq:x-kernel}, we have 
	\beq
	\begin{split}
		& k(P_r x, P_r \tilde{x}) \\
		& = \exp\left(-\frac{1}{h} (P_r x - P_r \tilde{x})^T (P_r x - P_r \tilde{x})\right)\\
		& = \exp\left(
		-\frac{1}{h} (w - \tilde{w})^T \Psi_r^T \Psi_r (w-\tilde{w})
		\right)\\
		& = \exp\left(
		-\frac{1}{h} ||w-\tilde{w}||_2^2
		\right)
	\end{split}
	\eeq
	where we used the relation $P_r x = \Psi_r w $ and $P_r \tilde{x} = \Psi_r \tilde{w} $ in the second equality and the orthonormality $\Psi_r^T \Psi_r = I$ in the generalized eigenvalue problem \eqref{eq:generalized-eigenproblem} in the third. Therefore, by definition \eqref{eq:w-kernel}, we have
	\beq
	k^r(w, \tilde{w}) = \kappa(x^r, \tilde{x}^r).
	\eeq
	Moreover, for the gradient of the kernel we have
	\beq
	\begin{split}
		\nabla_{x^r} \kappa(x^r, \tilde{x}^r) & = -\frac{2}{h} \kappa(x^r, \tilde{x}^r) (x^r - \tilde{x}^r)\\
		& = -\frac{2}{h} \kappa(x^r, \tilde{x}^r) \Psi_r (w - \tilde{w}).
	\end{split}
	\eeq
	On the other hand, we have 
	\beq
	\nabla_w k^r(w, \tilde{w}) = -\frac{2}{h} k^r(w, \tilde{w}) (w-\tilde{w}),
	\eeq
	which yields 
	\beq
	\nabla_w k^r(w, \tilde{w}) = \Psi_r^T \nabla_{x^r} \kappa(x^r, \tilde{x}^r).
	\eeq
	For the posterior $\eta$ defined in \eqref{eq:pi-p}, we have 
	\beq
	\nabla_{x^r} \log \eta(x^r) = \frac{\nabla_{x^r}(g(x^r)\eta_0(x^r))}{g(x^r)\eta_0(x^r)},
	\eeq
	while for the posterior $\pi$ defined in \eqref{eq:w-posterior}, we have 
	\beq
	\nabla_w \log \pi(w) = \frac{\nabla_w(g(\Psi_r w) \pi_0(w))}{g(\Psi_r w) \pi_0(w)}.
	\eeq
	By chain rule, it is straightforward to see that 
	\beq\label{eq:grad-g-star}
	\nabla_w g(\Psi_r w) = \Psi_r^T \nabla_{x^r} g(x^r).
	\eeq
	Under assumption $\pi_0(w) = p_0(P_r x )$ in Theorem \ref{thm:pSVGD}, and $p_0(P_r x )= \eta_0(x^r)$ by definition \eqref{eq:pi-p}, we have 
	\beq\label{eq:grad-p-0}
	\nabla_w \pi_0(w) = \Psi_r^T \nabla_{x^r} \eta_0(x^r).
	\eeq
	Therefore, combining \eqref{eq:grad-g-star} and \eqref{eq:grad-p-0}, we have 
	\beq
	\nabla_w \log \pi(w) = \Psi^T_r \nabla_{x^r} \log \eta(x^r).
	\eeq
	To this end, we obtain the equivalence of the Stein operators
	\beq
	\cA_{\pi}k^r(w, \tilde{w}) = \Psi_r^T \cA_\eta \kappa(x^r, \tilde{x}^r)
	\eeq
	for $x^r = \Psi_r w$ and $\tilde{x}^r = \Psi_r \tilde{w}$. Since the prior densities $\eta_0(x^r) = \pi_0(w)$, we have the equivalence
	\beq
	\bE_{w\sim \pi_0} [\cA_{\pi} k^r(w, \tilde{w})] = \Psi_r^T \bE_{x^r \sim \eta_0}[\cA_{\eta} \kappa(x^r, \tilde{x}^r)],
	\eeq
	which concludes the equivalence of the transport map \eqref{eq:projection-equivalence}
	by $w = \Psi^T_r x^r$ with the same $\epsilon$ at $\ell=0$. Moreover, by induction we have 
	\beq
	T^r_\ell(w^\ell) = \Psi_r^T T_\ell(P_r x^\ell),
	\eeq
	which concludes.
	
\end{proof}

\section{A linear inference problem}
\label{sec:linear-inference}

This example is presented to test the accuracy of the proposed algorithm with analytically given posterior distribution for a linear inference problem.
We consider a linear parameter-to-observable map $A: \bR^d \to \bR^s$, which is given by
\beq\label{eq:linear-map}
A x = O \circ B x,
\eeq
where $B: x \to u $ is a linear discrete solution map of the diffusion reaction equation ($\Delta$ is the Laplace operator)
\beq\label{eq:linear-pde}
-\Delta \mathrm{u} + \mathrm{u} = \mathrm{x}, \quad \text{ in } (0, 1),
\eeq
with boundary condition $\mathrm{u}(0) = 0$ and $\mathrm{u}(1) = 1$, which is solved by a finite element method. The continuous parameter $\mathrm{x}$ and solution $\mathrm{u}$ are discretized by finite elements with piecewise linear elements in a uniform mesh of size $d$. $x \in \bR^d$ and $u \in \bR^d$ are the nodal values of $\mathrm{x}$ and $\mathrm{u}$.
The parameter $\mathrm{x}$ is assumed to follow a Gaussian distribution $\cN(0, \cC)$ with covariance $\cC = (-0.1 \Delta + I)^{-1}$, which leads to a Gaussian parameter $x \sim \cN(0, \Sigma_x)$, with covariance $\Sigma_x \in \bR^{d \times d}$ as a discretization of $\cC$. 

$O:\bR^d \to \bR^s$ in \eqref{eq:linear-map} is an observation map that take $s$ components of $u$ that are equally distributed in $(0, 1)$. For $s = 15$, we have $O u = (u(1/16), \dots, u(15/16))^T$. We assume an additive $1\%$ Gaussian noise $\xi \sim \cN(0, \Sigma_\xi)$ with $\Sigma_\xi = \sigma^2 I$ and $\sigma = \max (|O u|)/100$ for data 
\beq
y = A x + \xi, 
\eeq
then the likelihood function is given by 
\beq
f(x) = \exp\left(- \frac{1}{2}||y-Ax||_{\Sigma_\xi^{-1}}^2\right). 
\eeq
Because of the linearity of the inference problem, the posterior of $x$ is also Gaussian $\cN(x_{\text{MAP}}, \Sigma_y)$ with the MAP point $x_{\text{MAP}} = \Sigma_y A^T\Sigma_\xi^{-1} y$ and covariance 
\beq\label{eq:post-cov}
\Sigma_y = (A^T \Sigma_\xi^{-1} A + \Sigma_x^{-1})^{-1}.
\eeq

\begin{figure}[!htb]
	\centering
	\includegraphics[width=0.7\textwidth]{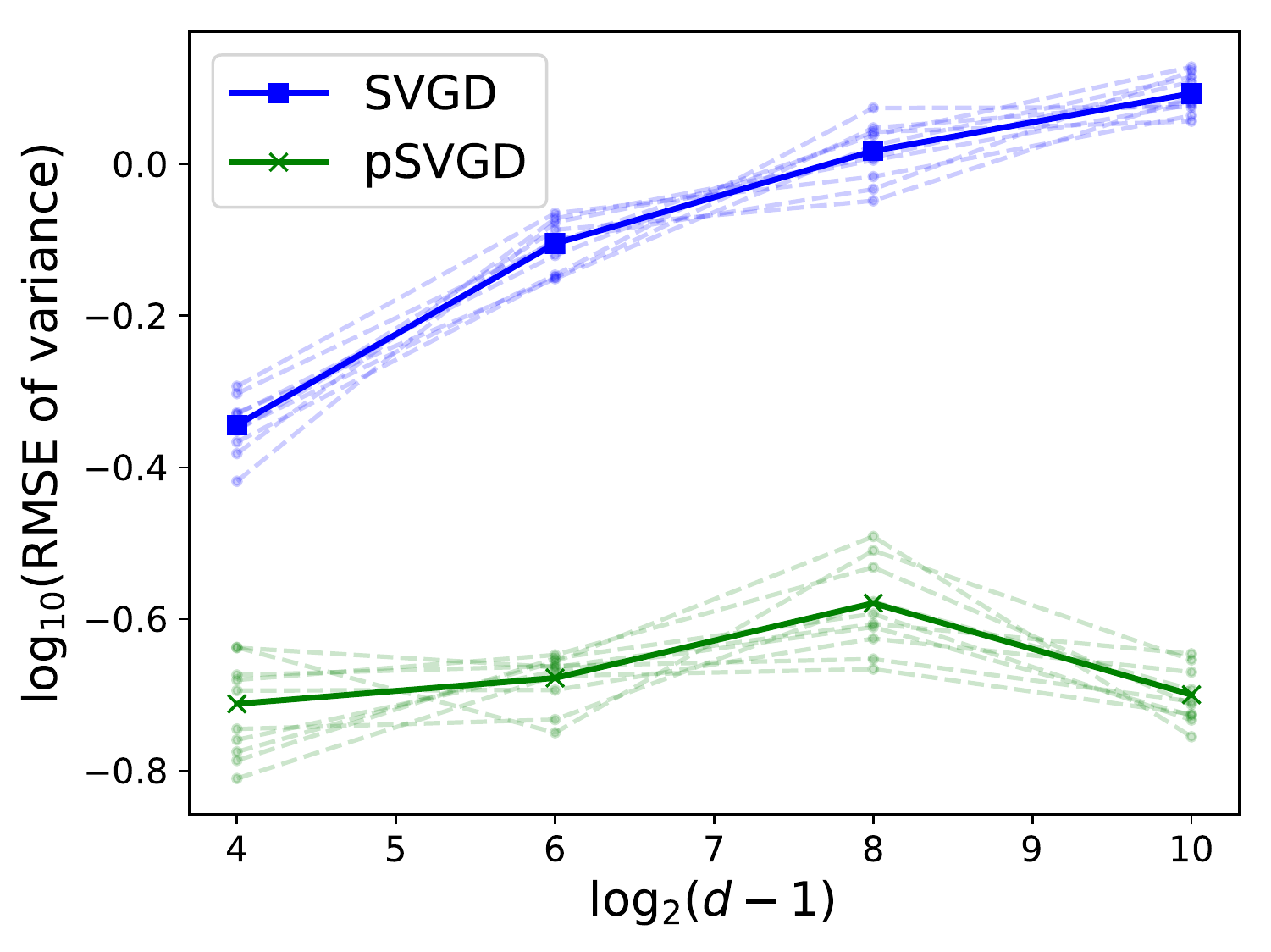}
	\vskip -0.1in
	\caption{RMSE of pointwise sample variance in $L_2$-norm, with 256 samples, SVGD and pSVGD both terminated at $\ell = 200$ iterations, parameter dimension $d = 2^n+1$, with $n = 4, 6, 8, 10$.}
	\label{fig:accuracy-linear}
\end{figure}

We run SVGD and pSVGD (projection with $r = 8$ basis functions and $\lambda_9 < 10^{-4}$) with 256 samples and 200 iterations for different dimensions, both using line search to seek the step size $\epsilon_\ell$. The RMSE (of 10 trials and their average) of the samples variances compared to the ground truth \eqref{eq:post-cov} are shown in Figure \ref{fig:accuracy-linear}, 
which indicates that SVGD deteriorates with increasing dimension while pSVGD performs well for all dimensions.

\section{Application in COVID-19}
\label{sec:covid19-inference}

\begin{figure}[!htb]
	\centering
	\includegraphics[width=1.0\textwidth]{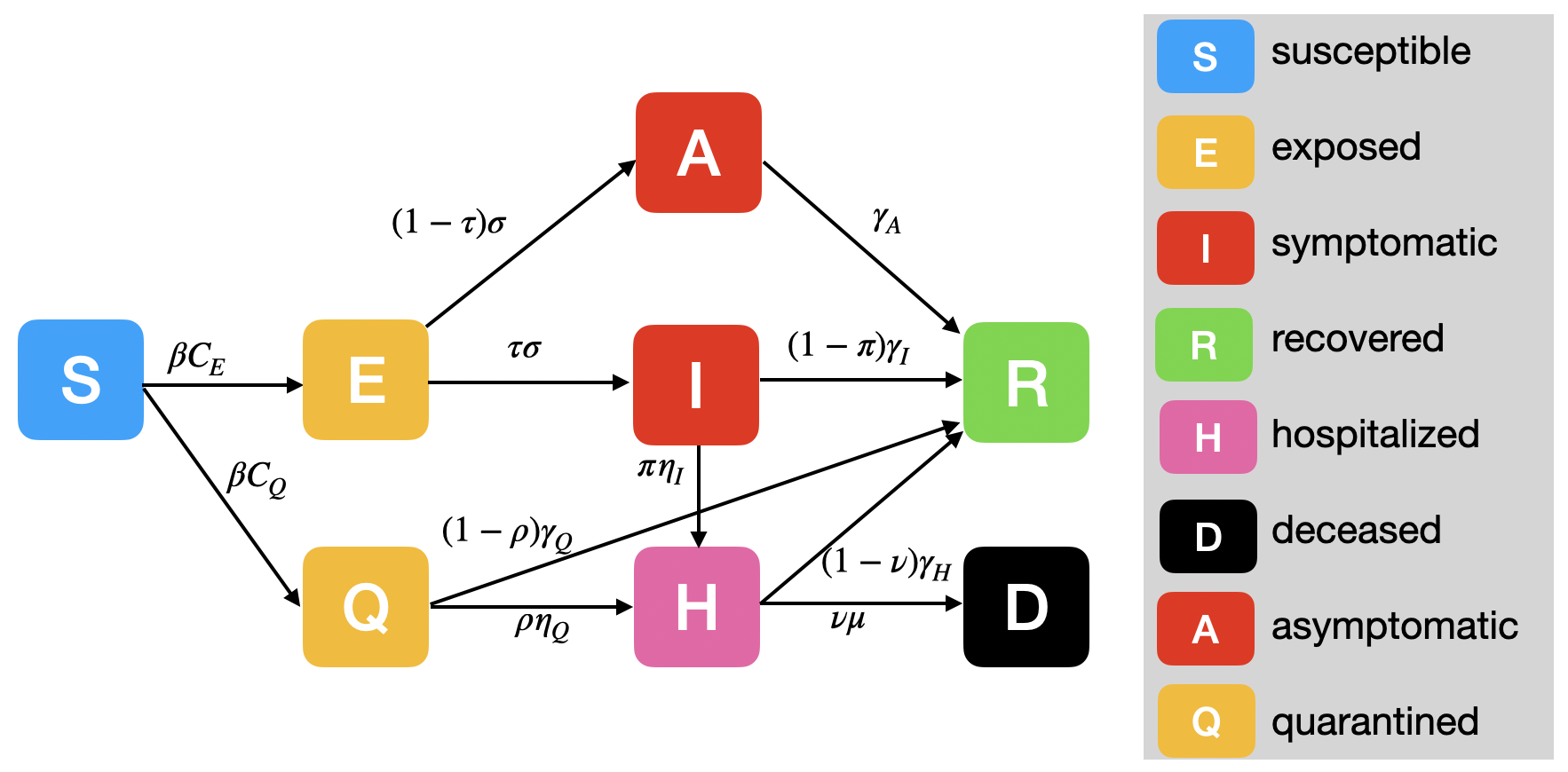}
	\caption{Sketch of a compartmental epidemic model with 8 compartments for modeling of transmission and outcome of infectious diseases such as COVID-19. }\label{fig:SEIR diagram}
\end{figure}

Social distancing has played a key role in flattening the curve of the spread of COVID-19. In this example, we apply pSVGD to infer a time-dependent parameter that describes the contact reduction effect of social distancing given observation data. We consider a compartmental model with 8 compartments for the modeling of the transmission and outcome of COVID-19, as illustrated by the diagram in Figure \ref{fig:SEIR diagram}, which is given by the system of ordinary differential equations

\begin{equation}\label{eq:forward}
\begin{split}
C_E(t)				 & = (1-\alpha(t))(1-q)  \, \frac{I(t)}{N} + (1-\alpha(t)) \frac{ A(t)}{N},
\\
C_Q(t)				 & = (1-\alpha(t))q  \, \frac{I(t)}{N},
\\
\frac{dS(t)}{dt} & = - \beta C_E(t)S(t) - \beta C_Q(t)S(t),
\\
\frac{dE(t)}{dt} & =  \beta C_E(t)S(t)  - \tau \sigma E(t) - (1-\tau) \sigma E(t),
\\
\frac{dQ(t)}{dt} & = \beta C_Q(t) S(t) - \rho \eta_Q Q(t) - (1-\rho) \gamma_Q Q(t),
\\
\frac{dA(t)}{dt} & = (1-\tau)\sigma E(t) - \gamma_A A(t) ,
\\
\frac{dI(t)}{dt} & = \tau \sigma E(t) - \pi \eta_I I(t) - (1-\pi) \gamma_I I(t)  ,
\\
\frac{dH(t)}{dt} & = \pi \eta_I I(t) + \rho \eta_Q Q(t) - \nu \mu H(t) - (1-\nu) \gamma_H H(t) ,
\\
\frac{dR(t)}{dt} & = \gamma_A A(t) + (1-\pi) \gamma_I I(t) + (1-\nu) \gamma_H H(t) + (1-\rho) \gamma_Q Q(t),
\\
\frac{dD(t)}{dt} & = \nu \mu H(t).
\end{split}
\end{equation}
In this model, $\alpha(t) \in [0, 1]$ represents the effective contact reduction of social distancing at time $t$, i.e., the percentage of reduced contact compared to the status without social distancing, which is the time-dependent parameter we infer. Briefly on the other parameters, $N$ is the total population for a given region, $\beta$ is a transmission rate, $q$ is quarantined rate, $\sigma$ is latency rate, $\eta_I, \eta_Q$ are hospitalized rates, $\gamma_A, \gamma_I, \gamma_Q, \gamma_H$ are recovery rates, $\mu$ is deceased rate, $\tau, \rho, \pi, \nu$ are the proportions of cases going from $E$ to $I$, $Q$ to $H$, $I$ to $H$, and $H$ to $D$. We assume these parameters are scalar and do not change over time. We use the number of hospitalized cases $H$ (7 days' moving average) in New York available in \url{https://github.com/COVID19Tracking}
as the observation data and assume that the observation noise is i.i.d. $N(0, 1)$ for the logarithm of the data to create the likelihood function. 

\begin{figure}[!htb]
	\centering
	\includegraphics[width=0.55\linewidth]{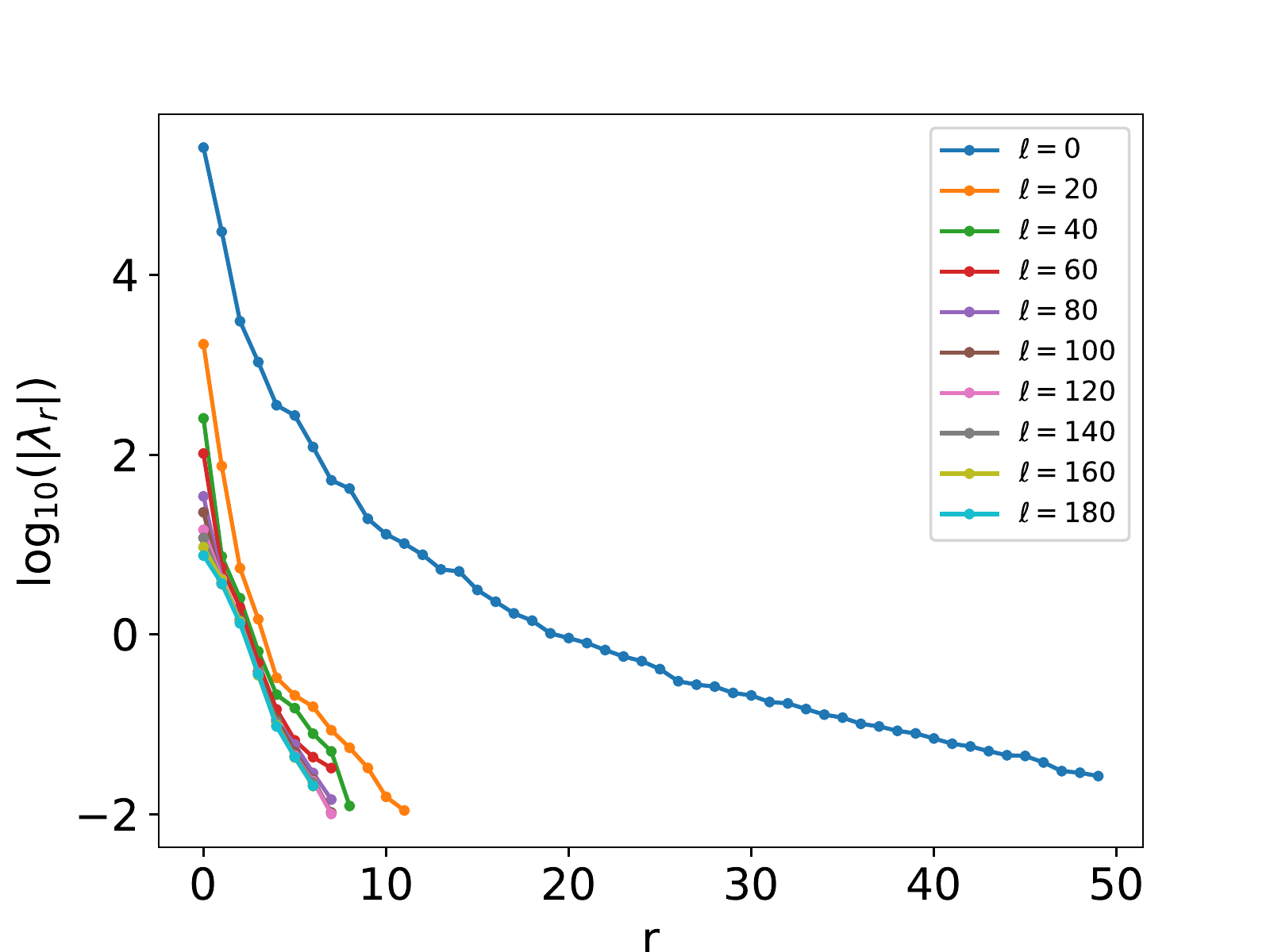}
	\makebox[\linewidth][c]{				
		\includegraphics[width=0.55\linewidth]{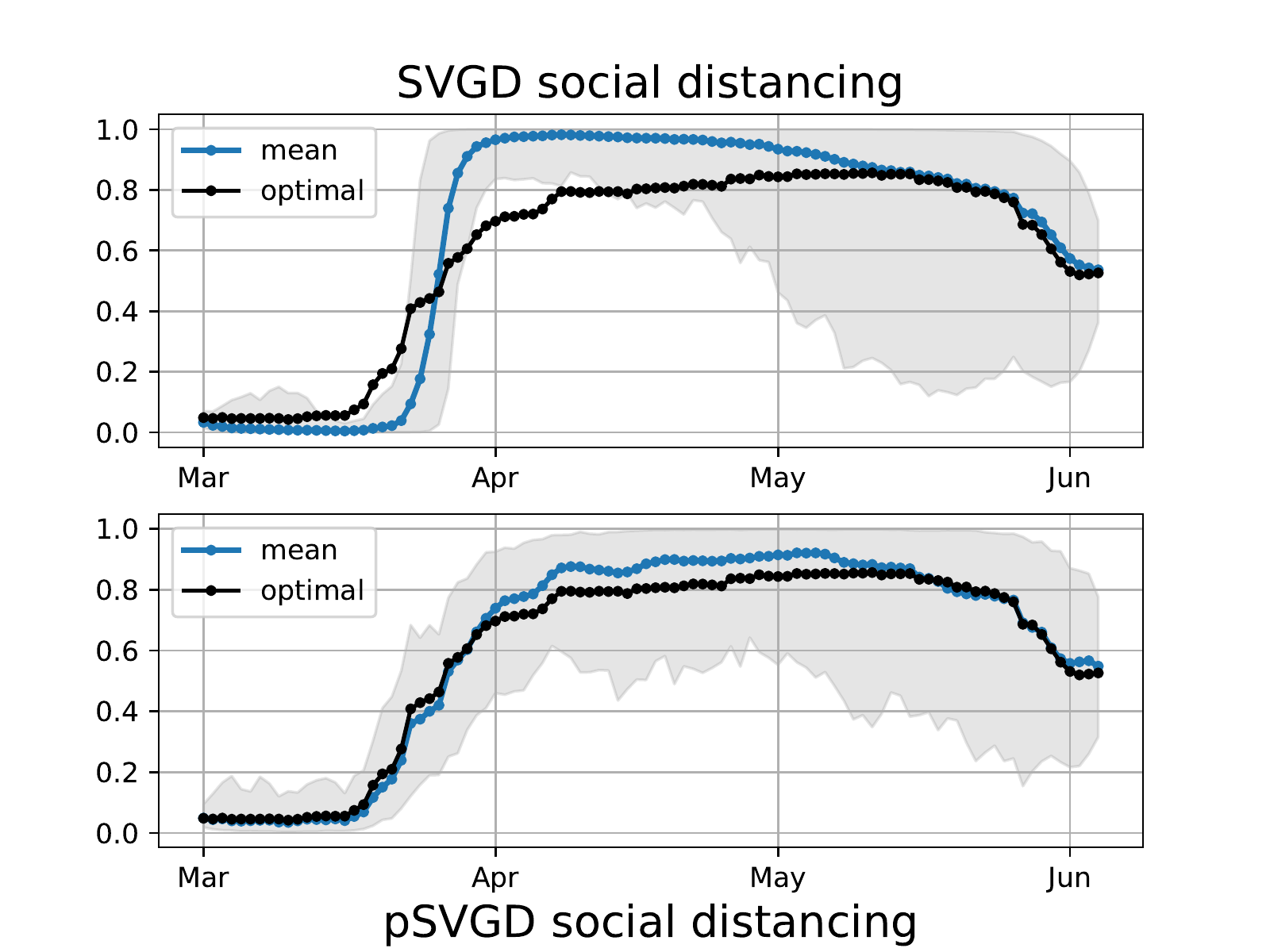}\hskip -0.3in
		\includegraphics[width=0.55\linewidth]{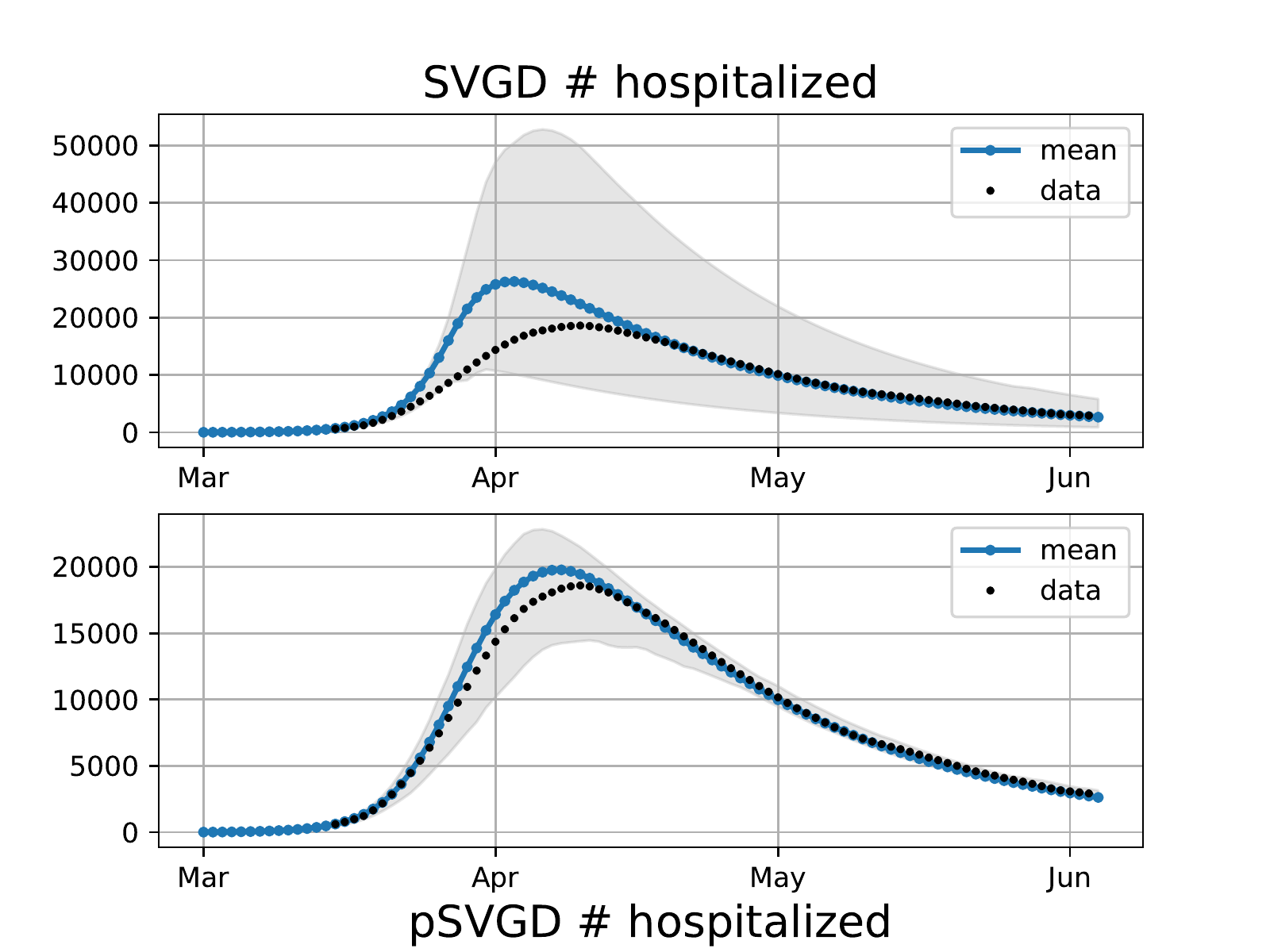}\hskip -0.3in
	}
	\caption{Top: Decay of the eigenvalues of \eqref{eq:generalized-eigenproblem} at different iteration numbers $\ell$. Bottom-left: samples of the reduction factor $\alpha$ of social distancing, posterior mean, 90\% confidence interval in shadow, deterministic optimal $\hat{\alpha}$. Bottom-right: the number of hospitalized cases, posterior mean, 90\% confidence interval in shadow,  and reported data for New York. The results are at iteration $\ell = 200$.}\label{fig:covid19}
	\vskip -0.1in
\end{figure}

First, we deterministically infer all these parameters by solving an optimization problem to minimize the misfit between the logarithm of the predicted number of hospitalized cases and that of the observed number. Then we freeze all the parameters at their optimal values except for $\alpha(t)$. We assume that 
$$
\alpha(t) = \frac{1}{2}(\tanh(x(t))+1)
$$ 
is a stochastic process with Gaussian process $x(t) \sim \cN(\hat{x}(t), \cC)$, where $\hat{x}(t) = \text{arctanh}(2\hat{\alpha}(t) - 1)$ at the deterministic optimal values of the social distancing $\hat{\alpha}(t)$ obtained from the deterministic optimization problem, $\cC = -(\delta \triangle_t)^{-1}$ with Laplacian operator $\triangle_t$ and a scaling parameter $\delta = 10$. After discretization in time with step of one day over 96 days, we obtain a discrete parameter $x = (x_1, \dots, x_d)$ of dimension $d = 96$. We run pSVGD and SVGD with line search, 128 samples with 8 samples in each of 16 processor cores, update the bases for pSVGD every 10 iterations for a total of 200 iterations. The results are shown in Figure \ref{fig:covid19}. We can observe a fast decay of eigenvalues and a small number of intrinsic dimension. The bottom two figures display the samples, their mean, and 90\% confidence interval of the reduction factor $\alpha$ for social distancing and the number of hospitalized cases by SVGD (top) and pSVGD(bottom). We can observe that pSVGD provides much more accurate prediction of the data (the number of hospitalized cases) with tighter confidence interval than SVGD from the right of Figure \ref{fig:covid19}. Meanwhile, the mean of the pSVGD samples of the reduction factor $\alpha$ of social distancing is much closer to the deterministic optimal value than that of SVGD, with 90\% confidence interval of pSVGD covering the deterministic optimal value, while that of SVGD not. The mean of the reduction factor $\alpha$ for SVGD is nearly 1 for a period of time, which corresponds to complete shutdown without any transmission, which is not realistic.

%
%
%
%
%
%
%
%
%
%
%
%
%
%
%
%
%
%
%
%
%
%


\end{document}